\documentclass[10pt,journal,compsoc]{IEEEtran}

\usepackage{algorithm}
\usepackage{algorithmic}
\usepackage{graphicx}
\usepackage{subfigure}
\usepackage{booktabs}
\usepackage[colorlinks,anchorcolor=blue]{hyperref}
\usepackage[nosumlimits,nonamelimits]{amsmath}
\usepackage{enumitem}
\usepackage{amssymb}
\usepackage{mathtools}
\usepackage{amsthm}
\usepackage{thmtools, thm-restate}
\usepackage{ragged2e}  % justifying

\usepackage[table]{xcolor}
\usepackage{cite}
\DeclareMathOperator{\rank}{rk}
\DeclareMathOperator{\spn}{span}

\usepackage{multirow}
\usepackage{colortbl}
\newcommand{\ts}{\textsuperscript}
\newcommand{\etal}{et al.}
\newtheorem{theorem}{Theorem}

\newtheorem{definition}{Definition}

\graphicspath{{figures/}}

\begin{document}

\title{Geometric Understanding of Discriminability and Transferability for Visual Domain Adaptation}

\author{You-Wei Luo,~Chuan-Xian~Ren,~Xiao-Lin~Xu,~and~Qingshan Liu% <-this % stops a space
\IEEEcompsocitemizethanks{\IEEEcompsocthanksitem Y.W. Luo and C.X. Ren are with the School of Mathematics, Sun Yat-Sen University, Guangzhou 510275, China. X.L. Xu is with the School of Statistics and Mathematics, Guangdong University of Finance and Economics, Guangzhou 510320, China. Q.S. Liu is with the School of Computer Science, Nanjing University of Posts and Telecommunications, Nanjing 210023, China. C.X. Ren is the corresponding author (email: rchuanx@mail.sysu.edu.cn).
\protect\\
\IEEEcompsocthanksitem This work is supported in part by National Natural Science Foundation of China (Grant No. 62376291, U21B2044), in part by Guangdong Basic and Applied Basic Research Foundation (2023B1515020004), in part by
Science and Technology Program of Guangzhou (2024A04J6413), in part by Guangdong Province Key Laboratory of Computational Science at Sun Yat-sen University (2020B1212060032), and in part by the Fundamental Research Funds for the Central Universities, Sun Yat-sen University.}% <-this % stops an unwanted space
%\thanks{Manuscript received Oct. 31, 2021.}
}

\markboth{IEEE TRANSACTIONS ON PATTERN ANALYSIS AND MACHINE INTELLIGENCE}%
{Shell \MakeLowercase{\textit{et al.}}: Bare Demo of IEEEtran.cls for Computer Society Journals}


\IEEEtitleabstractindextext{
\begin{abstract}
\justifying
To overcome the restriction of identical distribution assumption, invariant representation learning for unsupervised domain adaptation (UDA) has made significant advances in computer vision and pattern recognition communities. In UDA scenario, the training and test data belong to different domains while the task model is learned to be invariant. Recently, empirical connections between transferability and discriminability have received increasing attention, which is the key to understand the invariant representations. However, theoretical study of these abilities and in-depth analysis of the learned feature structures are unexplored yet. In this work, we systematically analyze the essentials of transferability and discriminability from the geometric perspective. Our theoretical results provide insights into understanding the \textit{co-regularization} relation and prove the \textit{possibility} of learning these abilities. From methodology aspect, the abilities are formulated as geometric properties between domain/cluster subspaces (i.e., orthogonality and equivalence) and characterized as the relation between the norms/ranks of multiple matrices. Two optimization-friendly learning principles are derived, which also ensure some intuitive explanations. Moreover, a feasible range for the co-regularization parameters is deduced to balance the learning of geometric structures. Based on the theoretical results, a geometry-oriented model is proposed for enhancing the transferability and discriminability via nuclear norm optimization. Extensive experiment results validate the effectiveness of the proposed model in empirical applications, and verify that the geometric abilities can be sufficiently learned in the derived feasible range.
\end{abstract}

}

\maketitle

\IEEEdisplaynontitleabstractindextext
\IEEEpeerreviewmaketitle

\IEEEraisesectionheading{\section{Introduction}\label{sec:introduction}}

\IEEEPARstart{W}{ith} the availability of large-scale datasets and development of deep learning, computer vision and pattern recognition tasks, such as image classification and object detection, have been rapidly advanced. However, standard learning models with the \textit{identical distribution assumption} have clear limitations in the changing real-world environments \cite{quinonero2008dataset,pan2009survey}. Specifically, when training and test data belong to different domains (e.g., synthetic image domain and real image domain), applying a model trained on the source domain directly to a new environment (target domain) often results in evident performance degradation. Such a problem setting is also known as unsupervised domain adaptation (UDA) \cite{courty2016optimal,long2018transferable,li2022generalized,li2021maximum,tang2021towards,luo2022unsupervised,gu2022unsupervised,xia2022maximum}. To overcome the limitations of classical learning framework and explore generalization models for UDA, an advanced learning paradigm called invariant representation learning \cite{zhao2019learning,li2021learning,nguyen2021domain,kirchmeyer2022mapping} has received increasing attention recently, which requires the prior knowledge can be transferred across different domains with the learned representations.

In general, two types of frameworks are frequently used to provide mathematically definition. 1) \textit{Statistical framework}: the domains are formulated as probability distributions. Several theoretical results \cite{ben2010theory,zhao2019learning,kirchmeyer2022mapping} are deduced from the perspective of statistics, which proves that the risk estimation on the source domain can be a good approximation for target domain if the domain discrepancy is minimized. Following this line, statistical models are used to learn invariant representations by minimizing the statistical divergence between domains \cite{long2018transferable,ganin2016domain,courty2016optimal,li2020enhanced}. 2) \textit{Geometric framework}: the domains are considered as vector spaces or manifolds, and the learning principles are usually based on matrix norm and subspace basis. The adaptation process implies the alignment of the basis vectors between different domains \cite{fernando2013unsupervised,gong2014learning,luo2022unsupervised,chen2021representation}. Compared with the statistical framework, the geometric framework usually ensures more intuitive explanations from the perspectives of linear algebra or Riemannian manifold. Besides, the geometry-based learning principles are usually connected with the optimization-friendly proxies, e.g., norms \cite{qiu2015learning}. But, for both statistical and geometric frameworks, theoretical understanding of the learned representations is insufficient currently.

\begin{figure}[t]
    \centering
    \includegraphics[width=0.93\linewidth,trim=20 36 45 20,clip]{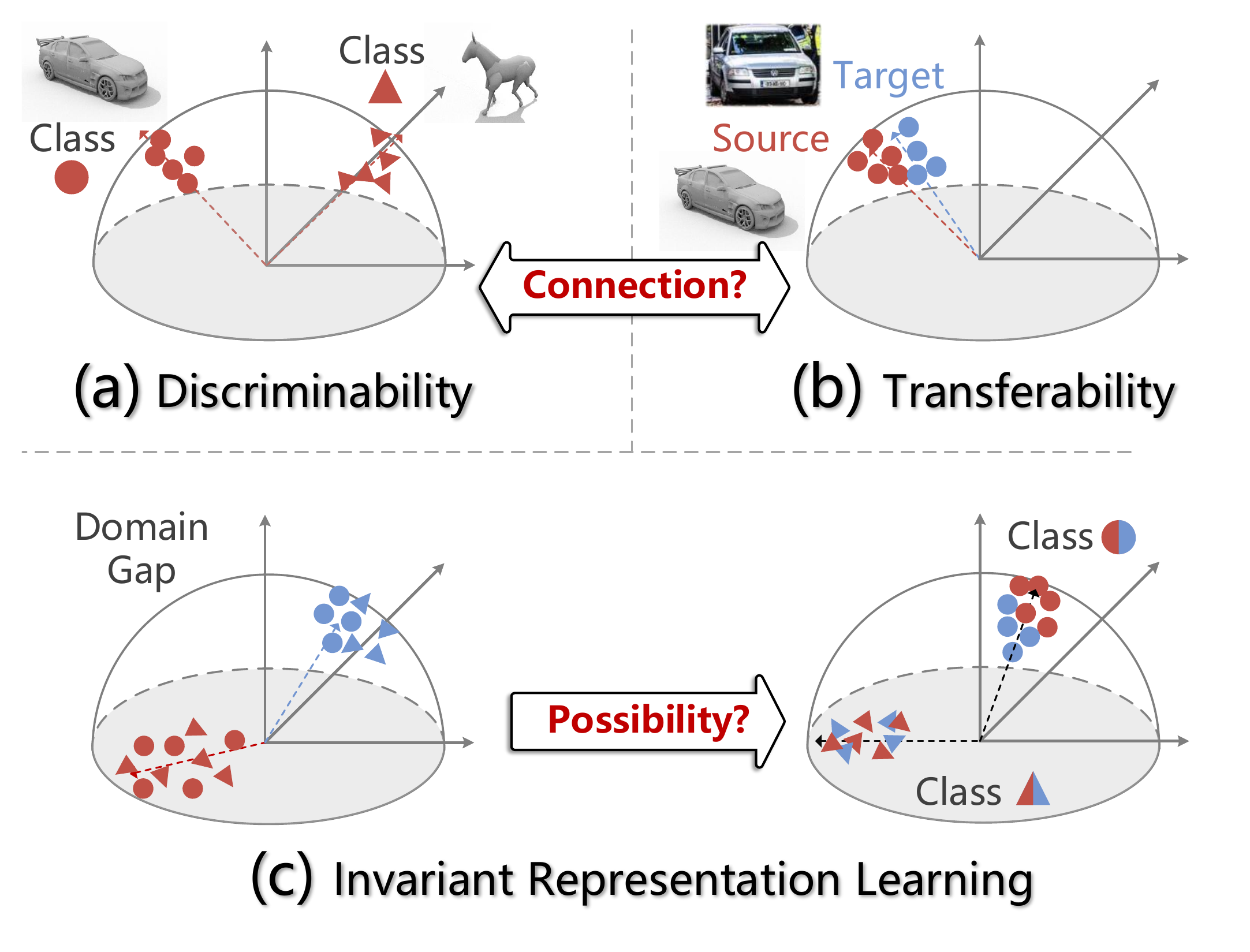}
    \caption{Problem illustration. \textbf{(a)-(b)}: A geometric view of discriminability and transferability. The bases in different classes are orthogonal, and the domains are linearly dependent. \textbf{(c)}: Invariant representation learning with simultaneously enhanced transferability and discriminability can reduce domain gap between changing environments. However, the possibility is unknown yet.}
    \label{fig:problem_illustration}
\end{figure}

Based on different mathematical frameworks, current UDA literature commonly focuses on two properties of the invariant representation model, i.e., transferability and discriminability \cite{chen2019transferability,xu2019larger}. Intuitively, transferability implies the confusion among the representations from different domains, and discriminability the identification among the representations from different classes. Generally, the abilities are characterized as the discrepancy between distributions in statistical framework, and as the geometric criteria (e.g., manifold metric, norm and rank) in geometric framework. Both statistical and geometric models have contributed a lot to learn transferability \cite{long2018transferable,ganin2016domain,courty2016optimal,luo2021conditional,stojanov2021domain}. While the discriminability of hidden representations is usually omitted and cannot be explicitly learned via the end-to-end deep learning model with entropy-based objectives \cite{yu2020learning}. Several works have been contributed to enhance discriminability of the hidden representations under standard learning scenarios \cite{lezama2018ole,yu2020learning,chan2021redunet}. Recently, by defining the transferability as principal angles and discriminability as linear discriminant analysis (LDA) value, Chen~\etal~\cite{chen2019transferability} empirically observed that UDA models focusing on transferability usually lead to degraded discriminability. Several works \cite{long2018conditional,li2020enhanced,cui2020towards} have tried to enhance the discriminability via different criteria while preserving the transferability of invariant representations.

Overall, the main limitations of current research are the lacks of systematic study on the properties of learned invariant representations, e.g., transferability and discriminability of representations. Besides, theoretical connections between these abilities are still unexplored yet. In this work, we aim to explore a rigorous theory and some intuitive interpretations for transferability and discriminability. Specifically, we focus on three problems. \textbf{1)} A theoretical understanding of transferability and discriminability. As shown in Figure~\ref{fig:problem_illustration}(a)-(b), we provide a geometric view of these abilities. The transferability and discriminability are mathematically defined as the relations between bases of vector spaces and measured by the norm/rank-based criteria, which connect the abstract abilities with intuitive geometric properties. \textbf{2)} The \textit{possibility} shown in Figure~\ref{fig:problem_illustration}(c). We are interested in the question ``is it possible to simultaneously learn these abilities?''. \textbf{3)} Effective principles for learning these abilities. To learn the desired geometric properties as shown in Figure~\ref{fig:problem_illustration}, the learning criteria are necessary and important. Based on the ranks/norms of multiple matrices, we propose principles for characterizing subspace bases learning these abilities. 

By dealing with the problems above, our main results provide following new insights. 1) Previous empirical conclusion is incomplete, since our theoretical results show that the discriminability and transferability are not strictly negatively correlated, and there is a balance state for them; 2) These two abilities are co-regularized and mutual beneficial, since we find that they prevent each other from the trivial solutions; 3) The existence of simultaneously maximized abilities are theoretically proved, the balance state can be explicitly learned under the guidance of theoretical results. Overall, our contributions can be summarized as follows.

\begin{itemize}
    \item A mathematical framework for understanding transferability and discriminability are developed from geometric aspects, where rigorous definitions, learning principles and their connections are studied. The results connect the abstract abilities with geometric properties and quantitative measures, which guarantee the interpretability and effectiveness.
    
    \item The \textit{co-regularization} relationship between transferability and discriminability is explored. The results show that there exists a balance state, where the abilities are mutually benefited. Based on the co-regularization understanding, the \textit{possibility} of maximizing two abilities simultaneously is proved.
    
    \item An analytic feasible region for balance state is derived, then a \textbf{g}eometry-\textbf{o}riented \textbf{a}bility \textbf{l}earning (GOAL) model is proposed. With nuclear norm proxy, GOAL is optimization-friendly. Extensive experiments validate the theoretical results and the empirical effectiveness.
\end{itemize}

The rest of this paper is organized as follows. Some related works on UDA and transferability/discriminability of invariant representations are reviewed in Section~\ref{sec:related_works}. In Section~\ref{sec:theoretical_analysis}, a systematic study on transferability and discriminability is presented. Based on the theoretical results, the GOAL model is proposed in Section~\ref{sec:GOAL_model}. The theory and model are experimentally validated in Section~\ref{sec:experiments}. Finally, we present conclusion and future directions in Section~\ref{sec:conclusion}.

\section{Related Works}\label{sec:related_works}

In this section, we briefly review some related works in UDA, and especially recent advances in transferability/discriminability of invariant representation learning.

\textbf{UDA.}
Under different math definitions, the mainstream UDA methods can be roughly summarized as follows.

\textit{Statistical framework} defines the domains as probability distributions and reduces the statistical discrepancy between domains to learn invariant model. Borgwardt~\etal~\cite{borgwardt2006integrating} use the kernel mean embedding to measure the maximum mean discrepancy (MMD) \cite{gretton2012kernel} between domains. Long~\etal~\cite{long2018transferable} extend the multi-kernel MMD metric to the deep learning model with better transferability by proposing deep adaptation network (DAN). Zhu~\etal~\cite{zhu2021deep} propose deep subdomain adaptation network (DSAN) to achieve class-wise domain alignment based on MMD. Li~\etal~\cite{li2021maximum} defines the maximum density discrepancy between domains and propose adversarial tight match (ATM) to learn domain density. Ganin~\etal~\cite{ganin2016domain} propose the domain-adversarial neural network (DANN) to formulate the adaptation problem as a two-player adversarial game, which theoretically minimizes the $\mathcal{H}$-divergence \cite{ben2010theory} between domains. Tang~\etal~\cite{tang2020discriminative} extend the adversarial adaptation by considering the discriminability. Gu~\etal~\cite{gu2022unsupervised} propose robust spherical domain adaptation (RSDA) to formulate adversarial training in spherical feature space, which improves the precision and robustness of pseudo-labels. Chen~\etal~\cite{chen2020adversarial} propose an adversarial training rule for domain adaptation (ALDA) to enhance the domain discriminator with confusion matrix. Wasserstein metric for UDA is proposed by Courty~\etal~\cite{courty2016optimal} based on the optimal transport (OT) theory. To study the OT problem in non-Euclidean space and derive the analytic solutions, Zhang~\etal~\cite{zhang2019optimal} and Luo~\etal~\cite{luo2021conditional} formulate the OT problems in reproducing kernel Hilbert space and prove the distribution embedding properties.

\textit{Geometric framework} defines the domains as vector spaces, graphs or Riemannian manifolds. Ghifary~\etal~\cite{ghifary2016scatter} propose the scatter component analysis to measure the domain discrepancy as the sample-level scatter values. Fernando~\etal~\cite{fernando2013unsupervised} employ the singular value decomposition (SVD) to align the singular vectors of domains, and then they prove the convergence of the alignment objective for empirical estimation. Gong~\etal~\cite{gong2016domain} formulate the domains as points on Grassmann manifold and generate the intermediate subspaces via the geodesic flow. Luo~\etal~\cite{luo2022unsupervised} consider the metrics on Grassmann and SPD manifolds, and then propose discriminative manifold propagation (DMP) to align the domains, where the theoretical error bounds for empirical estimations are also proved. Chen~\etal~\cite{chen2021representation} reduce the principal angles between domains to deal with UDA regression problem. Das~\etal~\cite{das2018unsupervised} and Xia~\etal~\cite{xia2020structure} match the graph relations of domains to learn geometric information. Yan~\etal~\cite{yan2018semi} employ Gromov-Wasserstein distance to match the graphs across domains. Though geometric framework usually ensures intuitive explanations of the learned representations, e.g., the angles between subspaces of domains and clusters, theoretical study on geometric properties for UDA and invariant representation learning is still unexplored.

\textbf{Transferability and Discriminability.}
Recently, transferability and discriminability of the invariant representation learning models gain rising attention \cite{chen2019transferability,zhao2019learning}, which is the key to understand the hidden representations learned from deep learning models. As most works focus on the transferability of representations, Chen~\etal~\cite{chen2019transferability} show that the local structures and label information for downstream tasks may be lost during learning transferability. Then, recent models are proposed to preserve the intrinsic information by enhancing the discriminability of representations. Li~\etal~\cite{li2020enhanced} propose enhanced transport distance (ETD) to incorporate the label information into OT problem. Long~\etal~\cite{long2018conditional} propose conditional domain adversarial networks to encode the entropy-aware weights into the adversarial training. Xiao~\etal~\cite{xiao2021dynamic} propose dynamic weighted learning (DWL) to balance the weights between domain alignment and class discrimination based on MMD and LDA values. Li~\etal~\cite{li2022generalized} propose generalized domain conditioned adaptation network (GDCAN) to preserve task-specific information via attention module. Besides, Norm-based criteria are also frequently used in learning these abilities. Chen~\etal~\cite{chen2019transferability} and Xu~\etal~\cite{xu2019larger} conclude that the representations with larger norms are more transferable while smaller norms more discriminative. Then Xu~\etal~\cite{xu2019larger} propose adaptive feature norm (AFN) to enhance transferability with $\ell_2$-norm maximization. Chen~\etal~\cite{chen2019transferability} propose batch spectral penalization (BSP) to enhance discriminability with spectral norm minimization. By considering the prediction matrix instead of representations, Cui~\etal~\cite{cui2020towards} propose batch nuclear-norm maximization (BNM) to ensure the diverse structure for predictions within batch data. Another fruitful line is self-training, which usually applies pseudo-labels to refine predictions and enhances the models' abilities, e.g., learning predictions with consistency \cite{sohn2020fixmatch,prabhu2021sentry,liu2021cycle}, class-balance \cite{zou2019confidence}, or identifiability \cite{jin2020minimum}. 

\textbf{Current Challenges.}
Overall, though the UDA models are extensively studied in recent years and many efforts have been made to enhance the invariant representations, a rigorous formulation and systematic study for the discriminability and transferability are unexplored yet. Besides, though recent works \cite{xu2019larger,chen2019transferability,cui2020towards} provide intuitive modeling for enhancing discriminability and transferability, the abilities are usually not considered under a unified framework, e.g., introducing spectrum-based criteria for discriminability and adversarial training for transferability. Thus, it is still unknown whether these two kinds of ability can be maximized simultaneously, i.e., lacking theoretical supports. Specifically, the common challenges can be summarized from several aspects.

\begin{itemize}[leftmargin=*]
    \item [1)] From the theoretical aspect, the mathematical definitions and sufficient learning principles for transferability and discriminability are unexplored in existing works. Besides, the strict trade-off understanding for these two abilities in existing works could be inappropriate, which implies that the enhanced discriminability will degrade the transferability and vice versa. Intuitively, the ideal framework should ensure a harmony relation, where the two abilities can be simultaneously maximized. 
    \item [2)] From the methodological aspect, existing works consider the norm of prediction/feature matrix individually, which cannot sufficiently characterize the relation between domains/clusters. Specifically, existing works generally characterize the geometric relation as linear dependency and independency, while the stronger properties, i.e., equivalence and orthogonality, are not well explored.
    \item [3)] From the aspect of effectiveness, existing works are generally insufficient to ensure transferability and discriminability, i.e., the objectives only satisfy necessary conditions. Moreover, the empirical feasibility of maximizing two abilities simultaneously under specific learning objectives is also unclear, which implies the possibility problem w.r.t. specific empirical model is still uncertain. 
\end{itemize}

In our work, we address these concerns from geometric perspectives and show the \textit{possibility} and \textit{principles} to enhance these abilities with theoretical guarantees and strong geometric properties. Specifically, we develop a unified geometric framework, which provides rigorous definition, effective objectives and reliable theoretical supports. Moreover, our results are consistent with the experiment observations. Therefore, our framework has advantages on both theoretical and methodological aspects.

\section{Theoretical Analysis}\label{sec:theoretical_analysis}

Generally, a labeled source domain $\mathcal{D}_s=\{(\mathbf{x}_i^s,\mathbf{y}_i^s)\}_{i=1}^{n^s}$ and an unlabeled target domain $\mathcal{D}_t=\{\mathbf{x}_i^t\}_{i=1}^{n^t}$ are given in UDA scenarios, where $\mathbf{x}$ is the feature vector and $\mathbf{y}$ is label with $k$ classes. Denote $\mathbf{X}^s\in \mathbb{R}^{D\times n^s}$, $\mathbf{X}^t\in \mathbb{R}^{D\times n^t}$ as data matrices, $\mathbf{Y}^s\in \mathbb{R}^{k\times n^s}$ as one-hot label matrix, $\mathbf{X}=[\mathbf{X}^s,\mathbf{X}^t]\in\mathbb{R}^{D\times n}$ as the concatenation of $\mathbf{X}^s$ and $\mathbf{X}^t$. Let $\mathbf{X}^s_i\in\mathbb{R}^{D\times n^s_i}$ and $\mathbf{X}_i=[\mathbf{X}^s_i,\mathbf{X}^t_i]\in\mathbb{R}^{D\times n_i}$ be the data matrices of $i$-th source class and the $i$-th class, respectively. Note that sample size satisfies that $n \!=\! n^s\!+\!n^t\! = \!n_1\!+\!n_2\!+\!\ldots\!+\!n_k$ and $n_i\!=\!n^s_i\!+\!n^t_i$. Generally, the superscript `$s/t$' means the domain, subscript `$i$' of capital (resp. lowercase) letter means the cluster (resp. sample). The operations $\rank(\mathbf{X})$ and $\|\mathbf{X}\|_*$ are rank and nuclear norm (also known as trace norm in Hilbert space) of $\mathbf{X}$, i.e., the sum of singular values of matrix $\mathbf{X}$. Without loss of generality, we use the ground-truth labels of target domain only for theoretical analysis, and replace them with pseudo-labels for empirical modeling in empirical modeling, which will be detailed in Section~\ref{subsec:alg}. \textbf{\textit{The proofs of theoretical results are provided in supplementary material}}.

\subsection{Geometric Views and Definitions}

Let $\mathcal{U}(\mathbf{X})=\{\mathbf{u}_1, \mathbf{u}_2,\ldots,\mathbf{u}_r\}$ be the orthonormal basis of column space of $\mathbf{X}$ where $r$ is rank of $\mathbf{X}$, e.g., the set of left singular vectors of $\mathbf{X}$ with non-zero singular values. Then the linear hull $\spn(\mathcal{U}(\mathbf{X}))$ is the column space of $\mathbf{X}$. With the orthonormal basis, we can characterize two given data matrices (e.g., domains or clusters) from a geometric view.
\begin{definition}
Let $\mathbf{A}$ and $\mathbf{B}$ be matrices of the same row dimensions, the bases $\mathcal{U}(\mathbf{A})$ and $\mathcal{U}(\mathbf{B})$ are called: \\
\textbf{(\romannumeral1)} \textit{equivalent}, denoted as $\mathcal{U}(\mathbf{A})\sim \mathcal{U}(\mathbf{B})$, if $\spn(\mathcal{U}(\mathbf{A})) = \spn(\mathcal{U}(\mathbf{B}))$. \\
\textbf{(\romannumeral2)} \textit{orthogonal}, denoted as $\mathcal{U}(\mathbf{A})\perp \mathcal{U}(\mathbf{B})$, if $\mathbf{a}\perp\mathbf{b}$ for any $\mathbf{a}\in\mathcal{U}(\mathbf{A})$ and $\mathbf{b}\in\mathcal{U}(\mathbf{B})$;
\end{definition}

Note that the orthogonality also admits $\spn(\mathcal{U}(\mathbf{A})) \cap \spn(\mathcal{U}(\mathbf{B})) = \varnothing$. The two properties above provide insights into understanding the transferability and discriminability with given data feature matrices. Specifically, we consider the learning problem for these properties on the projected subspaces, i.e., a transformation $g(\cdot): \mathbf{x}\mapsto\mathbf{z},~\mathbb{R}^{D}\rightarrow \mathbb{R}^{d}$. Note that the notations of $\mathbf{Z}$ are similar to those of $\mathbf{X}$. On the one hand, for learning diverse task knowledge, the subspaces (i.e., the bases) of different clusters are supposed to be orthogonal, i.e., disjoint, to achieve maximum discriminability. On the other hand, for transferring task knowledge and preserving intrinsic information, subspaces of the domains are supposed to be equivalent, i.e., overlapping, to achieve maximum transferability.
\begin{definition}
A representation transformation $g(\cdot)$ has: \\
\textbf{(\romannumeral1)} \textbf{maximum transferability} if class-wise domain bases are equivalent, i.e., $\mathcal{U}(\mathbf{Z}^s_i) \sim \mathcal{U}(\mathbf{Z}^t_i)$ for any $i\in \{1,2,\ldots,k\}$; \\
\textbf{(\romannumeral2)} \textbf{maximum discriminability} if cluster bases are orthogonal, i.e., $\mathcal{U}(\mathbf{Z}_i) \perp \mathcal{U}(\mathbf{Z}_j)$ for any $i\neq j$.
\end{definition}

\begin{figure}[t]
    \centering
    \includegraphics[width=0.90\linewidth,trim=10 2 10 3,clip]{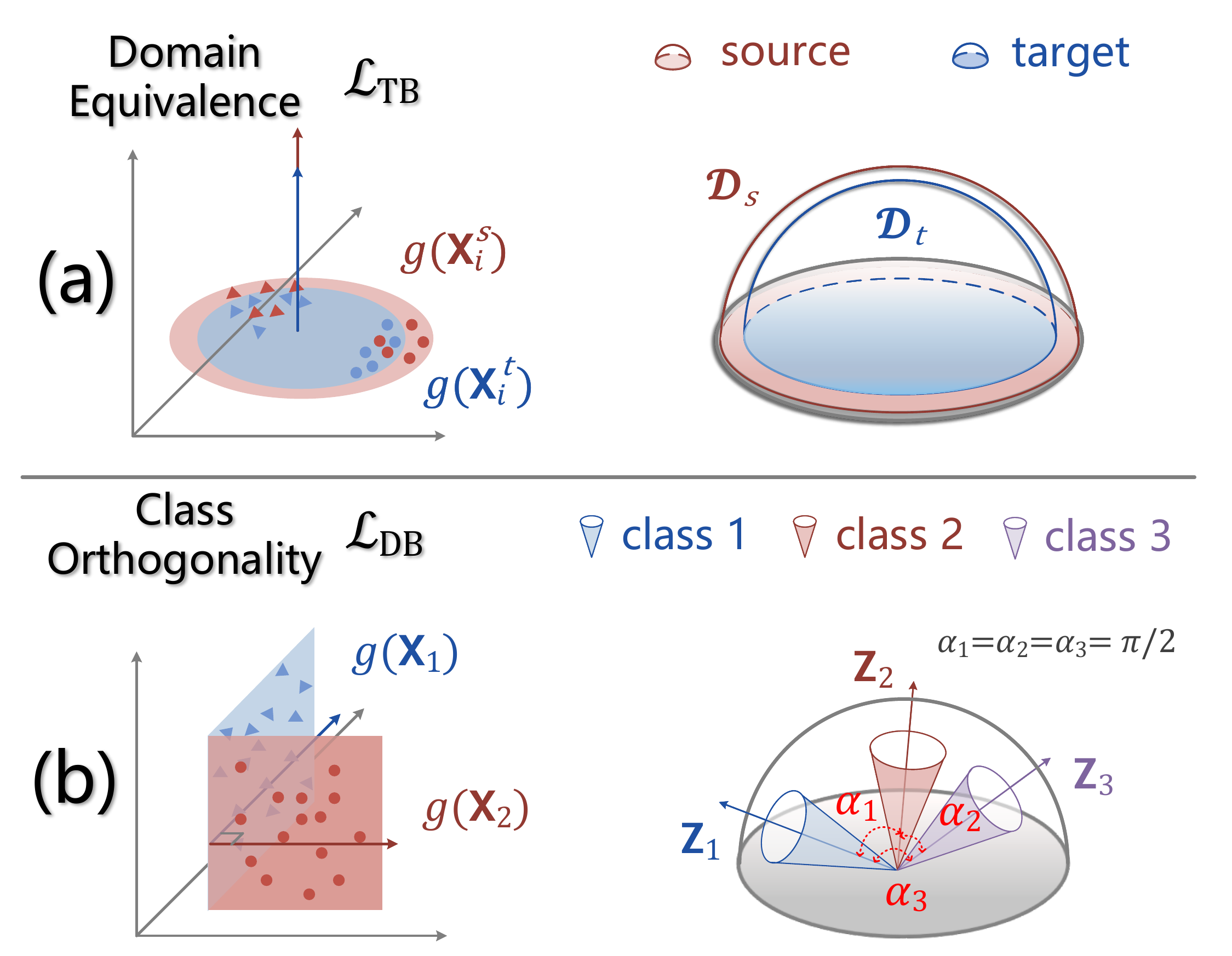} 
    \caption{A geometric view of transferability and discriminability. \textbf{(a)}: The left figure depicts that domains are overlapped to maximize the transferability; the right figure shows that the dimensions and dependence of subspaces are maximized when $\mathcal{L}_{\text{TB}}$ is optimized. \textbf{(b)}: The orthogonality between different clusters enhances the discriminability. Under $\mathcal{L}_{\text{DB}}$, the subspaces of different clusters are orthogonal which maximizes the angles between hyperplanes. }
    \label{fig:method_illustration}
\end{figure}

From the geometric perspectives, an intuitive interpretation for the definitions of transferability and discriminability is as shown in Figure~\ref{fig:method_illustration}. As Figure~\ref{fig:method_illustration}(a), the equivalence ensures the same clusters of different domains are in the same subspace/flat. As Figure~\ref{fig:method_illustration}(b), the orthogonality ensures that the clusters are distributed in different flats, then the representations of different clusters cannot be represented by each other. In the next, we will focus on the \textit{principles} of learning transferability and discriminability, and the \textit{possibility} of such a learning problem. The key is to find a quantitative measurement which can characterize the geometric properties.

\subsection{Motivation: Rank-Based Criteria}

The essential motivation is based on the rank of matrix, which is closely related to the geometry structure of the given data. It is known that the smaller the rank, the higher the correlation between the column vectors of (data) matrix. For the concatenation of domains, i.e., $[\mathbf{X}^s,\mathbf{X}^t]$, a linearly dependent structure for column space, where the source and target domains are correlated, is highly expected for transferability. In contrast, for the concatenation of clusters, i.e., $[\mathbf{X}_1,\mathbf{X}_2,\ldots,\mathbf{X}_k]$, blockwise independence is desired for learning identifiable cluster subspaces with enhanced discriminability.

To provide some insights into these geometric structures, we first review some preliminary results on the matrix rank. It is well-known that for matrices $\mathbf{A}$, $\mathbf{B}$, there are

\begin{equation*}
\rank ([\mathbf{A},\mathbf{B}])\leq \rank (\mathbf{A}) + \rank (\mathbf{B}),
\end{equation*}
where equality holds if and only if the bases of $\mathbf{A}$ and $\mathbf{B}$ are disjoint, i.e., $\mathcal{U}(\mathbf{A}) \cap \mathcal{U}(\mathbf{B}) = \varnothing$. It is clear that the inequity also holds for multiple matrices, i.e.,
\begin{equation}\label{eq:rank-upper-bound}
0\leq \sum_{i=1}^k \rank (\mathbf{A}_i)- \rank ([\mathbf{A}_1,\mathbf{A}_2,...,\mathbf{A}_k]).
\end{equation}
Besides, there is a lower bound of the rank of concatenation:
\begin{equation}\label{eq:rank-lower-bound}
   \max\{\rank (\mathbf{A}) , \rank (\mathbf{B})\} \leq  \rank ([\mathbf{A},\mathbf{B}]),
\end{equation}
where equality holds if and only if the basis of one matrix is a subset of the basis of another matrix, i.e., $\mathcal{U}(\mathbf{A})\subseteq \mathcal{U}(\mathbf{B})$ or $\mathcal{U}(\mathbf{B})\subseteq \mathcal{U}(\mathbf{A})$. Note that Eq.~\eqref{eq:rank-lower-bound} is equivalent to
\begin{equation}\label{eq:rank-lower-bound-2}
        \rank (\mathbf{A}) + \rank (\mathbf{B}) - \rank ([\mathbf{A},\mathbf{B}]) \leq \min\{\rank (\mathbf{A}) , \rank (\mathbf{B})\}.
\end{equation}
The analysis above provides a motivation for learning transferability and discriminability via rank. Specifically, for the transformed features $\mathbf{z}$, the concatenation of clusters $[\mathbf{Z}_1,\mathbf{Z}_2,\ldots,\mathbf{Z}_k]$ is expected to reach the equality in Eq.~\eqref{eq:rank-upper-bound}, where the bases of different classes are disjoint; the concatenation of domains $[\mathbf{Z}_i^s,\mathbf{Z}_i^t]$ is expected to reach the equality in Eq.~\eqref{eq:rank-lower-bound} at class-level, where the subspaces of domains are overlapped. Then one domain can be represented by the basis of another. Now, the rank-based learning criteria for \textbf{T}ransfera\textbf{B}ility and \textbf{D}iscrimina\textbf{B}ility can be formulated as
\begin{align}
    \mathop{\arg \max}_{g(\cdot)} ~ \mathcal{J}_{\text{TB}} =& \sum_{i=1}^k \rank (\mathbf{Z}_i^s) + \rank (\mathbf{Z}_i^t) - \rank (\mathbf{Z}_i), \label{eq:rank_TB_criteria} \\
    \mathop{\arg \min}_{g(\cdot)} ~ \mathcal{J}_{\text{DB}} =&   \sum_{i=1}^k \rank (\mathbf{Z}_i) - \rank (\mathbf{Z}). \label{eq:rank_DB_criteria}
\end{align}

Generally, the rank-based principles are not affected by the relation between feature dimension and sample size, since rank is essentially characterized by the basis of matrix. Further, if a low-dimensional embedding assumption is considered (i.e., feature dimension is smaller than sample size), which is reasonable since the principles are applied to the embedding representations $\mathbf{Z}$, we can conclude some appealing results as follows.

\begin{restatable}{remark}{RankCor}
    \label{rem:rank-based-TBDB}
    A transformation $g(\cdot)$ has: \\
    \textbf{(\romannumeral1)} \textbf{maximum transferability} if it is the solver of Eq.~\eqref{eq:rank_TB_criteria}; \\
    \textbf{(\romannumeral2)} \textbf{maximum discriminability} only if it is the solver of Eq.~\eqref{eq:rank_DB_criteria}.
\end{restatable}

Remark~\ref{rem:rank-based-TBDB} can be straightforwardly deduced from Eq.~\eqref{eq:rank-upper-bound} and Eq.~\eqref{eq:rank-lower-bound-2}. Note that though the rank-based criteria provide a feasible solution to learn geometric properties, the rank-based optimization is still an NP-hard problem, e.g., it cannot be directly applied as a plug-and-play loss in neural networks. In the next, we focus on the approx for rank and prove the geometric properties for the approx-based criteria.

\begin{figure*}[t]
    \centering
    \includegraphics[width=0.85\linewidth,trim=30 32 30 25,clip]{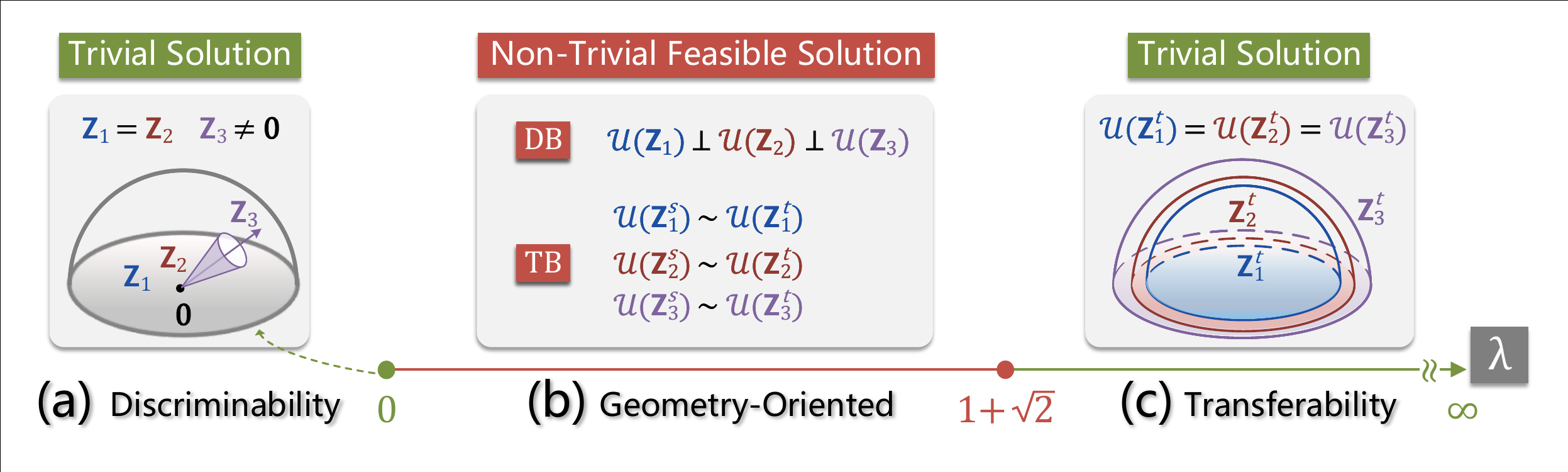}
    \caption{Illustration of the \textit{co-regularization}, including trade-off state (a)/(c) and balance state (b), between transferability and discriminability by varying the parameter $\lambda = \lambda_{\text{TB}}/\lambda_{\text{DB}}$. \textbf{(a)}: When $\lambda = 0$, only the discriminability objective is learned. The model may learn a trivial solution with most subspaces collapse to a point as zero element, i.e., Theorem~\ref{thm:TBDB_CoRegular}\textbf{(\romannumeral1)}. \textbf{(b)}: When $\lambda \in  (0,1+ \sqrt{2}]$, the co-regularization between transferability and discriminability reaches a balance. The model is possible to learn the geometric properties simultaneously as Theorem~\ref{thm:TBDB_CoRegular}\textbf{(\romannumeral3)}. \textbf{(c)}: When $\lambda \in  (1+ \sqrt{2},\infty]$, the model will over-learn the transferability. Then the discriminability will be degraded as Theorem~\ref{thm:TBDB_CoRegular}\textbf{(\romannumeral2)}. Especially, when $\lambda=\infty$, the model will learn transferability with totally overlapping clusters, i.e., the clusters are indistinguishable with minimum discriminability.}
    \label{fig:co_regularization_illustration}
    \vskip -0.10in
\end{figure*}

\subsection{Principles of Learning Geometric Properties}\label{subsec:norn_learning_principle}

Since nuclear norm is the convex envelop of rank over the unit ball of matrices, it is known to be a good proxy for rank optimization. Unfortunately, though nuclear norm can address the concerns on optimization, the geometric properties (i.e., orthogonality and equivalence) of nuclear norm are unclear. According to the motivation based on rank, the key is to study the relation between ``sum of norms'' (i.e., $\|\mathbf{A} \|+\|\mathbf{B}\|$) and  ``norm of concatenation'' (i.e., $\|[\mathbf{A},\mathbf{B}]\|$). With this goal in mind, we present the theoretical results for norm-based learning principles in the next.

\textbf{Transferability.}
A norm-based principle for learning transferability can be similarly formulated as Eq.~\eqref{eq:rank_TB_criteria}. However, the theoretical support for such a principle is unclear since the lower bound likes Eq.~\eqref{eq:rank-lower-bound-2} and the condition of the equality are unexplored for nuclear norm. In the next, we will address these problems and theoretically connect the norm-based objective with domain equivalence.

We first define norm-based principle for transferability:
\begin{equation}\label{eq:norm_TB_principle}
    \mathop{\arg \max}_{g(\cdot)} ~ \mathcal{L}_{\text{TB}} =\sum_{i=1}^k \| \mathbf{Z}_i^s \|_* + \| \mathbf{Z}_i^t \|_* - \| \mathbf{Z}_i \|_*.
\end{equation}
Here we are interested in two problems. 1) What is the upper bound of the maximization problem shown in Eq.~\eqref{eq:norm_TB_principle}? i.e., is this objective bounded and well-defined? 2) The connection between the optimal solution of Eq.~\eqref{eq:norm_TB_principle} and the defined maximum transferability, i.e., the relation between the optimal solution and domain equivalence.

For simplicity, we denote $\mathbf{A}\in \mathbb{R}^{d\times n}$, $\mathbf{B}\in \mathbb{R}^{d\times m}$ where row and column can be taken as the feature dimension and sample-size, respectively. As we consider model in a low-dimensional projective space, it is reasonable to assume $d\leq \min\{n,m\}$. We now present the transferability theorem.

\begin{restatable}{theorem}{TBthm}
    \label{thm:norm_TB_thm}
    Assume that $\| \mathbf{A} \|_\sigma \leq \alpha$, $\| \mathbf{B} \|_\sigma \leq \alpha$, where $\| \mathbf{A} \|_\sigma$ is the spectral norm, then \\
    \textbf{(\romannumeral1)} $\|\mathbf{A}\|_*+\|\mathbf{B}\|_* - \|[\mathbf{A},\mathbf{B}]\|_* \leq (2-\sqrt{2})\alpha d$; \\
    \textbf{(\romannumeral2)} the equality in (\romannumeral1) holds if and only if  $\|\mathbf{A}\|_*=\|\mathbf{B}\|_*=\alpha d$, which implies $\mathcal{U}(\mathbf{A}) \sim \mathcal{U}(\mathbf{B})$.
\end{restatable}

Theorem~\ref{thm:norm_TB_thm}\textbf{(\textit{\romannumeral1)}} shows that the transferability objective $\mathcal{L}_{\text{TB}}$ is upper-bounded by $(2-\sqrt{2})\alpha d$ in a ball of finite radius $\alpha$. Since nuclear norm is the convex envelop of rank over the unit ball \cite{fazel2001rank}, $\alpha$ is usually set as 1 in application. This result guarantees that the objective $\mathcal{L}_{\text{TB}}$ is bounded and optimization problem $\mathop{\max} \mathcal{L}_{\text{TB}}$ is well-defined.

Theorem~\ref{thm:norm_TB_thm}\textbf{(\textit{\romannumeral2)}} shows the interpretability of norm-based principle $\mathcal{L}_{\text{TB}}$ for learning transferable representations, and provides an insight into the norm-based learning. Specifically, source and target domains are projected into the equivalent subspace when $\mathcal{L}_{\text{TB}}$ reaches the maximum. Besides, the norms of representations are required to be maximized as $\|\mathbf{Z}^s_i\|_*=\|\mathbf{Z}^t_i\|_*=\alpha d$. It implies that the larger norm is sufficient to learn two linearly dependent domain subspaces with equivalent subspace bases. Furthermore, this result provides a theoretical justification to recent models on learning transferability \cite{xu2019larger,cui2020towards} which empirically show that the representations with larger norms are more preferable. Since the cross-domains features with maximized norm in a ball of finite radius will fill the subspace, the larger norm methods will also ensure the domain equivalence.

In fact, the overlap of subspace bases admits geometric interpretability for $\mathcal{L}_{\text{TB}}$. It means the spaces spanned by different domain bases are the same and the principal angles between them are 0. These properties ensure the task knowledge is preserved in the shared representation space. Therefore, the optimization of domain equivalence $\mathcal{L}_{\text{TB}}$ will project the clusters of different domains into the same subspaces as shown in Figure~\ref{fig:method_illustration}(a), where the knowledge is properly transferred.

\textbf{Discriminability.}
The norm-based principle for discriminability was studied by Qiu~\etal~\cite{qiu2015learning}, where a theoretical guarantee is provided as following theorem.

\begin{theorem}[\cite{qiu2015learning}]\label{thm:norm_DB_thm}
Let $\mathbf{A}$ and $\mathbf{B}$ be matrices with the same row dimensions, we have \\
\textbf{(\romannumeral1)} $\|[\mathbf{A},\mathbf{B}]\|_*\leq \|\mathbf{A}\|_*+\|\mathbf{B}\|_*$; \\
\textbf{(\romannumeral2)} the equality holds when $\mathcal{U}(\mathbf{A}) \perp \mathcal{U}(\mathbf{B})$.
\end{theorem}

Similarly, orthogonality means the maximum angle separation, that is, all principal angles between subspaces are equal to $\frac{\pi}{2}$. Note that rank-based criterion only ensures the bases are linearly independent, so it is not sufficient to learn orthogonality as depicted in Remark~\ref{rem:rank-based-TBDB}\textbf{(\romannumeral2)}. Generally, nuclear norm ensures a stronger geometric property than rank.

Based on these points, the norm-based principle for discriminability, as shown in Figure~\ref{fig:method_illustration}(b), is formulated as
\begin{equation}
    \label{eq:norm_DB_principle}
    \mathop{\arg \min}_{g(\cdot)} ~ \mathcal{L}_{\text{DB}} =   \sum_{i=1}^k \|\mathbf{Z}_i\|_*- \|\mathbf{Z}\|_*.
\end{equation}
According to Theorem~\ref{thm:norm_DB_thm}, the infimum of $\mathcal{L}_{\text{DB}}$ is reached when $\{\mathbf{Z}_i\}$ are orthogonal as shown in Figure~\ref{fig:method_illustration}(b). In this case, subspaces of different clusters will try to be orthogonal to each other and principal angles will be maximized.

\subsection{Possibility of Learning Geometric Properties}
Now we focus on the major problem, i.e., the possibility of learning transferability and discriminability simultaneously. By incorporating $\mathcal{L}_{\text{DB}}$ and $\mathcal{L}_{\text{TB}}$, we first define \textit{geometry-oriented} constraint for invariant representation learning as
\begin{equation}
\mathop{\arg \min}_{g(\cdot)} ~ \mathcal{L}_{\text{GO}} = \lambda_{\text{DB}}\mathcal{L}_{\text{DB}} - \lambda_{\text{TB}}\mathcal{L}_{\text{TB}},
\end{equation}
where $\lambda_{\text{TB}}$ and $\lambda_{\text{DB}}$ are non-negative parameters. Note that $\mathcal{L}_{\text{GO}}$ provides a unified view to understand orthogonality and equivalence in learning invariant representations. When $\lambda_{\text{TB}}=0$ or $\lambda_{\text{DB}}=0$, $\mathcal{L}_{\text{GO}}$ boils down to $\mathcal{L}_{\text{DB}}$ or $\mathcal{L}_{\text{TB}}$. Intuitively, the parameters are crucial for the theoretical properties of $\mathcal{L}_{\text{GO}}$, and a rigorous analysis is necessary.

\textbf{Trivial Solutions.}
Note that the choice of $\lambda_{\text{TB}}$ and $\lambda_{\text{DB}}$ can be taken as the mutual regularization between transferability and discriminability. Specifically, $\mathcal{L}_{\text{DB}}$ tries to learn compact subspaces for clusters but $\mathcal{L}_{\text{TB}}$ tends to learn a ``loose'' space with larger norm. We denote $\lambda = \frac{\lambda_{\text{TB}}}{\lambda_{\text{DB}}}$. On the one hand, if $\lambda \rightarrow 0$, the minimum of $\mathcal{L}_{\text{GO}}$ is dominated by $\mathcal{L}_{\text{DB}}$ which is 0. The model may learn a trivial solution such that $\mathcal{L}_{\text{GO}}\rightarrow 0$, e.g., $\mathbf{Z}_i\neq \mathbf{0}$ while other clusters are $\mathbf{0}$ as shown in Figure~\ref{fig:co_regularization_illustration}(a). On the other hand, if $\lambda \rightarrow \infty$, the minimum of $\mathcal{L}_{\text{GO}}$ is dominated by the maximum of transferability objective $\mathcal{L}_{\text{TB}}$ which is $(2-\sqrt{2})\alpha k d$. In this trivial case, all clusters share the same subspace and task discriminability is lost as shown in Figure~\ref{fig:co_regularization_illustration}(c).  

Therefore, $\mathcal{L}_{\text{GO}}$ cannot be taken as a simple combination of $\mathcal{L}_{\text{DB}}$ and $\mathcal{L}_{\text{TB}}$, since the principle may fail for some parameter settings of $\lambda_{\text{DB}}$ and $\lambda_{\text{TB}}$. Therefore, it is necessary to find theoretical guaranteed parameter settings for ensuring effectiveness of geometric principle $\mathcal{L}_{\text{GO}}$. In the next, we will show that if $\lambda_{\text{TB}}$ and $\lambda_{\text{DB}}$ are properly chosen, transferability and discriminability will serve as regularization for each other, and simultaneous learning of two abilities is possible.

\textbf{Co-regularization.}
Hopefully, the trivial solutions can be avoided by exploring the balance between $\lambda_{\text{TB}}$ and $\lambda_{\text{DB}}$. The key is to understand the trade-off and balance in $\mathcal{L}_{\text{GO}}$ when $\lambda$ moves from $0$ to $\infty$ as shown in Figure~\ref{fig:co_regularization_illustration}. For simplicity, denote $\tilde{\mathcal{L}}_{\text{GO}}= \frac{\mathcal{L}_{\text{GO}}}{\lambda_{\text{DB}}}$ and $\lambda = \frac{\lambda_{\text{TB}}}{\lambda_{\text{DB}}}$. We next show how to balance the parameters to achieve a favorable geometric structure via \textit{geometry-oriented} constraint. In this case, $\mathcal{L}_{\text{TB}}$ and $\mathcal{L}_{\text{DB}}$ are co-regularized.

\begin{restatable}{theorem}{TOthm}
\label{thm:TBDB_CoRegular}
Assume that $\| \mathbf{Z}^{s} \|_\sigma \leq \alpha$ and $\| \mathbf{Z}^{t} \|_\sigma \leq \alpha$.  \\
\textbf{(\romannumeral1)} \emph{(Discriminability)} If~ $\lambda =0$ ($\lambda_{\emph{DB}}\rightarrow \infty$), then $\tilde{\mathcal{L}}_{\emph{GO}} = \mathcal{L}_{\emph{DB}}$ and $\tilde{\mathcal{L}}_{\emph{GO}} \geq 0$. \\
\textbf{(\romannumeral2)} \emph{(Transferability)} If~ $1+ \sqrt{2} < \lambda \leq \infty$, then
\begin{equation*}
\tilde{\mathcal{L}}_{\emph{GO}} \geq \left[\left((\sqrt{2}-2)\lambda + \sqrt{2}\right)\sqrt{k}-\sqrt{2}\right] \alpha d ,
\end{equation*}
with equality when $\| \mathbf{Z}^{s}_i \|_* = \| \mathbf{Z}^{t}_i \|_* = \frac{1}{\sqrt{k}} \alpha d$ and $\mathcal{U}(\mathbf{Z}^s_i) \sim \mathcal{U}(\mathbf{Z}^t_i)$. It implies $\mathcal{U}(\mathbf{Z}_1), \ldots, \mathcal{U}(\mathbf{Z}_k)$ are not orthogonal. Especially, if $\lambda =\infty$ ($\lambda_{\emph{TB}}\rightarrow \infty$), then $\tilde{\mathcal{L}}_{\emph{GO}} = \lambda \mathcal{L}_{\emph{TB}}$. \\
\textbf{(\romannumeral3)} \emph{(Balance)} If~ $0< \lambda \leq 1+ \sqrt{2}$, then
\begin{equation*}
\tilde{\mathcal{L}}_{\emph{GO}} \geq (\sqrt{2}-2)\alpha \lambda d ,
\end{equation*}
with equality when $\mathcal{U}(\mathbf{Z}_1) \perp \mathcal{U}(\mathbf{Z}_2) \perp \ldots \perp \mathcal{U}(\mathbf{Z}_k)$ and $\mathcal{U}(\mathbf{Z}^s_i)\sim \mathcal{U}(\mathbf{Z}^t_i)$ for any $i\in \{1,2,\ldots,k\}$.
\end{restatable}

Theorem~\ref{thm:TBDB_CoRegular} splits $\tilde{\mathcal{L}}_{\text{GO}}$ into three cases as shown in Figure~\ref{fig:co_regularization_illustration}. In case \textbf{(\romannumeral1)} with $\lambda = 0$, $\tilde{\mathcal{L}}_{\text{GO}}$ boils down to $\mathcal{L}_{\text{DB}}$. Without the regularization of $\mathcal{L}_{\text{TB}}$, the clusters may collapse to a point as zero element, and trivial solutions like Figure~\ref{fig:co_regularization_illustration}(a) may occur. In case \textbf{(\romannumeral2)} with $\lambda \in  (1+ \sqrt{2},\infty]$, the domain equivalence $\mathcal{L}_{\text{TB}}$ is over-learned where the orthogonality among clusters is lost. This is because $\| \mathbf{Z}^{s/t}_i \|_* = \frac{1}{\sqrt{k}}\alpha d\Rightarrow \mathcal{L}_{\text{TB}}>0$, then the orthogonality condition will never be reached. That is to say, the enhancement of transferability will lead to a decreasing discriminability when $\lambda > 1+ \sqrt{2}$. Especially, Figure~\ref{fig:co_regularization_illustration}(c) show the scenario $\lambda=\infty$, where all clusters are equivalent and the discriminability is minimum. Thus, \textbf{(\romannumeral1)}-\textbf{(\romannumeral2)} implies the strict trade-off relation between two abilities. Case \textbf{(\romannumeral3)} suggests that a proper parameters $\lambda_{\text{TB}}$ and $\lambda_{\text{DB}}$ should satisfy that $\lambda \in  (0,1+ \sqrt{2}]$. In this case, class orthogonality $\mathcal{L}_{\text{DB}}$ provides a regularization which prevents $\mathcal{L}_{\text{TB}}$ from learning overlarge subspaces for clusters. Finally, an ideal geometric structure with simultaneously enhanced abilities is achieved as Figure~\ref{fig:co_regularization_illustration}(b), i.e., balance state.

\textbf{Possibility. }
Theorem~\ref{thm:TBDB_CoRegular}\textbf{(\romannumeral3)} implies that it is possible to learn the transferability and discriminability simultaneously from the perspective of geometry. Intuitively, it implies that a balanced relation between the two abilities is reached. Besides, Theorem~\ref{thm:TBDB_CoRegular} also implies that the transferability or discriminability will be overly learned and leads to degraded model. Therefore, our geometric-oriented principle is theoretically guaranteed to be effective in learning two abilities simultaneously. Note that the geometric properties are ensured for $\lambda \in  (0,1+ \sqrt{2}]$, while the learnability of $g(\cdot)$ may change with different $\lambda$ values which usually lead to different empirical results. We will verify this claim in numerical experiments, and empirically show that the parameters should also be ``balanced'' (i.e., $\lambda\approx 1$).

In conclusion, our theoretical results prove that such a co-regularization between transferability and discriminability indeed exists, which supports the recent observations in UDA and invariant representation learning \cite{xu2019larger,chen2019transferability}. Besides, the results provide appropriate ranges for balancing the learning of two abilities, which connect $\mathcal{L}_{\text{GO}}$ with desired geometric properties.

\section{A Geometry-Oriented Model}\label{sec:GOAL_model}

\subsection{Overall Model and Analysis}
\label{subsec:GOAL_model}
The overall GOAL model consists of two parts, i.e., the risk objective for task learning and the geometry-oriented constraint for invariant representation learning. Denote the classifier as $h(\cdot)$ and its probabilistic predictions by $\hat{\mathbf{y}}=h(\mathbf{z})\in\mathbb{R}^k$, where $\hat{y}_i\geq 0 $ and $\sum_{i=1}^k \hat{y}_i = 1$. Let $\hat{\mathbf{Y}}\in\mathbb{R}^{k\times n}$ be the prediction matrix. We will try to learn the representation transformation $g(\cdot)$ and classifier $h(\cdot)$ with geometric properties. The goal of UDA is learning a task model $\hat{\mathbf{y}}=h(g(\mathbf{x}))$ such that the target prediction $\hat{\mathbf{Y}}^t$ is close to the unknown ground-truth $\mathbf{Y}^t$.

\textbf{Task Learning.}
A common way to learn a task knowledge model is empirical risk minimization. Here we apply this rule with entropy-based cost functions to both domains. First, the cross-entropy objective is applied to the source domain with ground-truth label for supervised learning, i.e.,
\begin{equation*}
\mbox{\small$\displaystyle
\mathop{\arg \min}_{g(\cdot), ~h(\cdot)}~ \mathcal{L}^s_{\text{E}} = \sum_{j=1}^{n^s}\sum_{i=1}^k - \mathbf{Y}_{ij}^s \log \hat{\mathbf{Y}}_{ij}^s.
$}
\end{equation*}
For the target domain without labels, the entropy minimization is applied for unsupervised learning, which can also be considered as a regularization, i.e.,
\begin{equation*}
    \mbox{\small$\displaystyle
    \mathop{\arg \min}_{g(\cdot), ~h(\cdot)}~ \mathcal{L}^t_{\text{E}} = \sum_{j=1}^{n^t}\sum_{i=1}^k - \hat{\mathbf{Y}}_{ij}^t \log \hat{\mathbf{Y}}_{ij}^t.
    $}
    \end{equation*}
It preserves the classification knowledge and minimizes the uncertainty of prediction.

\textbf{Invariant Representation Learning.}
By integrating the geometry-oriented constraint, the overall objective is
\begin{equation}\label{eq:overall_objective}
\mathop{\arg \min}_{g(\cdot), ~h(\cdot)}~ \mathcal{L}_{\text{GOAL}} = \mathcal{L}_{\text{Task}}(g, h) + \mathcal{L}_{\text{GO}}(g),
\end{equation}
where $\mathcal{L}_{\text{Task}}=\mathcal{L}^s_{\text{E}} + \lambda_t \mathcal{L}^t_{\text{E}}$ is the task learning objective. Note that though task learning loss $\mathcal{L}_{\text{Task}}(g, h)$ is a signal to improve the identifiability of classifier, the transformation $g(\cdot)$ learned from $\mathcal{L}_{\text{Task}}(g, h)$ is usually a black-box model, and the properties of hidden representations are unknown. Recent study \cite{lezama2018ole,yu2020learning,chan2021redunet} also points out that the discriminability of representations usually cannot be explicitly learned by the entropy-based objectives with end-to-end optimization. Since our definitions for abilities focus on the representations, the task learning objective cannot be rigorously considered as discriminability principle. Hopefully, in GOAL, the geometric regularization $\mathcal{L}_{\text{GO}}(g)$ provides interpretability for the representations extracted by $g(\cdot)$. Specifically, the geometry-oriented constraint encourages that the representations of the same class are as compact as possible, and the subspaces of different classes are orthogonal. Besides, subspaces of different domains are enlarged as much as possible to learn the domain equivalence, which is compatible with the diverse clusters learning, i.e., maximum norm with full-rank.

\textbf{Complexity Analysis.}
The computational complexity of the learning principle Eq.~\eqref{eq:overall_objective} consists of two parts. For the entropy-based task learning model $\mathcal{L}_{\text{Task}}$, it can be computed in linear time with complexity $\mathcal{O}(kn)$. For the geometry-oriented constraint $\mathcal{L}_{\text{GO}}$, the main complexity consists of the SVDs of cluster and domain matrices. Specifically, note that the complexity of SVD of a $d\times n$ matrix is $\mathcal{O}(\min\{d^2n,dn^2\})$, then the complexity of $\mathcal{L}_{\text{TB}}$ and $\mathcal{L}_{\text{DB}}$ are $\mathcal{O}(d^2n)$. Overall, GOAL has linear complexity $\mathcal{O}(kn+d^2n)$.

\textbf{Pseudo-Label Analysis.}
For modeling with pseudo-label, the major concerns focus on the uncertain incorrect predictions and number of selected labels. 1) For uncertainty, our norm-based principles are generally robust to the incorrect pseudo-labels due to the optimal approximation property of SVD. Specifically, for the samples with incorrect pseudo-labels, they will be assigned to the incorrect clusters in learning procedure. Note that nuclear norm is characterized by the importance of basis (i.e., singular values). Then for the basis vectors in $\mathcal{U}$, the vectors of correct samples usually have larger contribution (i.e., large singular values and low approximation error) to reconstruct the cluster matrix compared with incorrectly assigned samples. Therefore, the norm-based principles are mainly dominated the correct samples in clusters. Furthermore, when predictions are overly confident on one class, it usually implies this class is head class with large sample size. Thus, the dominant singular values are still determined by the correct samples. 2) For label selection, rank-based objectives and norm-based objectives can be still effective when there are less selected pseudo-labels. Specifically, since both rank and nuclear norm are measured by numbers and importance of basis vectors, the objectives mainly depend on the number of `representative' samples, i.e., the fewest samples that form a basis. Note the `representative' samples are relatively easy to be identified since they characterize the essentials of clusters (i.e., basis), which implies that they usually have high prediction confidences when the predictions are not confident on many samples. Therefore, the `representative' samples are likely to be selected from all samples, then the geometry-based objectives can be effective when the model are not well-trained, e.g., initial stage.

\begin{algorithm}[t]
    \caption{GOAL for Invariant Representation Learning}
    \label{alg:GOALforUDA}
    \begin{algorithmic}[1]
    \REQUIRE Source data $\mathcal{D}_s$, Target data $\mathcal{D}_t$, Warm-up epochs $T_{\text{Warm}}$, GOAL epochs $T_{\text{Adapt}}$, Learning rate $\beta$, geometric principle parameters $\lambda_{\text{DB}}$ and $\lambda_{\text{TB}}$;\\
    \ENSURE Representation transformation $g(\cdot)$, classifier $h(\cdot)$;
    \STATE Initialize the network parameters $\Theta=\{\Theta_g,\Theta_h\}$; \\
    \% \textit{\textbf{Warm-Up Stage}}
    \FOR {$iter=1,2,...,T_{\text{Warm}}$}
    \STATE Forward propagate $\{\mathbf{x}_i^s\}_{i=1}^{n^s}$ as $\mathbf{z}=g(\mathbf{x})$, $\hat{\mathbf{y}}=h(\mathbf{z})$; \\
    \STATE Compute $\mathcal{L}_{\text{Warm}}=\mathcal{L}^s_{\text{E}}+\lambda_{\text{TB}}\mathcal{L}_{\text{TB}}+\lambda_{\text{DB}}\mathcal{L}_{\text{DB}}$, where $\mathcal{L}_{\text{TB}}$ aligns the domains globally and $\mathcal{L}_{\text{DB}}$ only uses $\mathcal{D}_s$;
    \STATE Update: $\Theta\leftarrow\Theta-\beta\triangledown\mathcal{L}_{\text{Warm}}(\Theta)$;
    \ENDFOR\\
    \% \textit{\textbf{GOAL Learning Stage}}
    \FOR {$iter=1,2,...,T_{\text{Adapt}}$}
    \STATE Forward propagate $\{\mathbf{x}_i^s\}_{i=1}^{n^s}$ and $\{\mathbf{x}_i^t\}_{i=1}^{n^t}$ and compute the pseudo-labels $\bar{\mathbf{Y}}^t$;
    \STATE Compute $\mathcal{L}_{\text{GOAL}} = \mathcal{L}_{\text{Task}} + \lambda_{\text{TB}}\mathcal{L}_{\text{TB}}+\lambda_{\text{DB}}\mathcal{L}_{\text{DB}}$ as Eq.~\eqref{eq:overall_objective};\\
    \STATE Update: $\Theta\leftarrow\Theta-\beta\triangledown\mathcal{L}_{\text{GOAL}}(\Theta)$;
    \ENDFOR
    \end{algorithmic}
\end{algorithm}

\subsection{Algorithm}
\label{subsec:alg}
As neural network (NN) generally ensures better capacity for learning, we instantiate $g(\cdot)$ and $h(\cdot)$ as NNs. The NN-based model is optimized with batch gradient descent as shown in Algorithm~\ref{alg:GOALforUDA}. Since the geometry-oriented constraint $\mathcal{L}_{\text{GO}}$ requires labels of the target data, we use the classifier to assign pseudo-labels $\bar{\mathbf{Y}}^t\in\mathbb{R}^{k\times n^t}$ for the target domain. Specifically, $\bar{\mathbf{Y}}_{ij}^t = 1$ if $i=\mathop{\arg \max}_l \hat{\mathbf{Y}}_{lj}^t$. To reduce the risk brought by the error prediction, we set a threshold $\tau$ to select pseudo-labels with confidences larger than $\tau$.

To reduce the uncertainty of pseudo-label, we design a multi-stage pipeline. As shown in Algorithm~\ref{alg:GOALforUDA}, there are two components, namely, warm-up and geometry structure learning. In the first stage, the source data with ground-truth labels and target data without labels are used. At this point, the source classification objective $\mathcal{L}^s_{\text{E}}$, class orthogonal objective $\mathcal{L}_{\text{DB}}$ and global $\mathcal{L}_{\text{TB}}$ are optimized, where only source data are applied to $\mathcal{L}_{\text{DB}}$ and $\mathcal{L}_{\text{TB}}$ are computed at domain-level. In the geometry learning stage, the entire GOAL model is optimized according to Eq.~\eqref{eq:overall_objective}, where the source data with ground-truth labels and target data with pseudo-labels are used.

\section{Numerical Experiments}\label{sec:experiments}

Numerical experiments are conducted on four visual UDA benchmarks for classification-oriented transfer as follows.

\textbf{Office-31}~\cite{saenko2010adapting} contains 4,110 images of 31 categories collected from 3 domains: Amazon (A), Web camera (W) and Digital SLR camera (D).

\textbf{Image-CLEF}~\cite{caputo2014ImageCLEF} consists of 3 domains: Caltech-256 (C), ImageNet (I) and Pascal-VOC (P). There are 12 categories and each class contains 50 images. 

\textbf{VisDA-2017}~\cite{peng2017visda} contains about 152k synthetic images from Synthetic (S) domain and 55k real-world images from Real (R) domain. Each domain contains 12 object categories.

\textbf{DomainNet}~\cite{peng2019moment} contains about 596k images from 6 domains: Clipart (clp), Infograph (inf), Painting (pnt), Quickdraw (qdr), Real (rel) and Sketch (skt). Each domain contains 345 categories

\textbf{Office-Home}~\cite{venkateswara2017deep} contains object images from 4 domains: Artistic (Ar), Clipart (Cl), Product (Pr) and Real-World (Rw). Each domain contains 65 object categories, and they amount to around 15,500 images.

\textbf{Implementation Details}.
The representation transformation $g(\cdot)$ consists of the ResNet backbone \cite{he2016deep} with 2048-dimensional output and two fully-connected layers that project the deep features as $\mathbb{R}^{2048}\rightarrow \mathbb{R}^{1024} \rightarrow \mathbb{R}^{512}$; the classifier $h(\cdot):\mathbb{R}^{512}\rightarrow \mathbb{R}^{k}$ is single fully-connected layer. The warm-up stage is implemented on the whole network with batch-size of 32; after warm-up, the deep backbone is fixed and only two fully-connected layers in $g$ and classifier $h$ are optimized on GOAL learning stage, which permits larger batch-size for better approximation in geometric structure learning, e.g., 16k on VisDA-2017 and DomainNet. The overall optimization is implemented with ADAM optimizer. \textit{More details for numerical implementation and additional experiments are provided in supplementary material.}

\subsection{Comparison with SOTA}

%\subsubsection{Comparison}

We compare GOAL with several recently developed SOTA UDA methods: DAN~\cite{long2018transferable}, DANN~\cite{ganin2016domain}, MCD~\cite{saito2018maximum}, SAFN~\cite{xu2019larger}, CRST~\cite{zou2019confidence}, BSP~\cite{chen2019transferability}, MDD~\cite{zhang2019bridging}, BNM~\cite{cui2020towards}, FixMatch~\cite{sohn2020fixmatch}, MCC~\cite{jin2020minimum}, DWL~\cite{xiao2021dynamic}, DSAN~\cite{zhu2021deep}, ATM~\cite{li2021maximum}, GDCAN~\cite{li2022generalized}, SCDA~\cite{li2021semantic}, DMP~\cite{luo2022unsupervised} and RSDA~\cite{gu2022unsupervised}.

\begin{table}[t]
    \centering
    \caption{Classification accuracies (\%) on \textbf{Office-31} and \textbf{Image-CLEF} (ResNet-50).}
    \label{tab:compare_31_CLEF}
    \renewcommand{\tabcolsep}{0.18pc} 
    \begin{tabular}{lccccccc}
    \toprule
    \textbf{Office-31} & A$\rightarrow$W           & A$\rightarrow$D           & W$\rightarrow$A           & W$\rightarrow$D          & D$\rightarrow$A           & D$\rightarrow$W          & Avg.          \\
    \hline
    Source \cite{he2016deep} & 68.4 & 68.9 & 60.7 & 99.3 & 62.5 & 96.7 & 76.1 \\
    DAN    \cite{long2018transferable}   & 84.2 & 87.3 & 65.2 & \textbf{100.0} & 66.9 & 98.4 & 83.7    \\
    DANN  \cite{ganin2016domain}   & 91.4 & 83.6 & 70.4 & \textbf{100.0} & 73.3 & 97.9 & 86.1    \\
    MCD \cite{saito2018maximum} & 90.4 & 87.3 & 67.6 & \textbf{100.0} & 68.3 & 98.5 & 85.4   \\
    SAFN \cite{xu2019larger} & 94.0 & 94.4 & 71.1 & \textbf{100.0} & 72.9 & 98.9 & 88.6    \\
    BSP+CDAN \cite{chen2019transferability} & 93.3 & 93.0 & 72.6 & \textbf{100.0} & 73.6 & 98.2 & 88.5 \\
    MDD \cite{zhang2019bridging} & 95.6 & 94.4 & 72.2 & \textbf{100.0} & 76.6 & 98.6 & 89.6   \\
    BNM \cite{cui2020towards} & 91.5 & 90.3 & 71.6 & \textbf{100.0} & 70.9 & 98.5 & 87.1 \\
    FixMatch \cite{sohn2020fixmatch} & 86.4 & 95.4 & 68.1 & 100.0 & 70.0 & 98.2 & 86.4 \\
    MCC \cite{jin2020minimum} & 94.1 & 95.6 & 74.2 & 99.8 & 75.5 & 98.4 & 89.6 \\
    DSAN \cite{zhu2021deep} & 93.6 & 90.2 & 74.8 & \textbf{100.0} & 73.5 & 98.3 & 88.4 \\
    ATM \cite{li2021maximum} & 95.7 & {96.4} & 73.5 & \textbf{100.0} & 74.1 & \textbf{99.3} & 89.8 \\
    GDCAN \cite{li2022generalized} & 94.8 & 93.6 & 74.4 & \textbf{100.0} & 76.9 & 98.2 & 89.7 \\
    SCDA+CDAN \cite{li2021semantic} & 94.7 & 95.4 & 76.0 & \textbf{100.0} & 77.1 & 98.7 & 90.3\\
    DMP   \cite{luo2022unsupervised}   & 93.0          & 91.0          & 70.2          & \textbf{100.0}         & 71.4          & 99.0         & 87.4          \\
    RSDA-MSTN \cite{gu2022unsupervised} & \textbf{95.9} & 96.1 & 78.2 & \textbf{100.0} & 77.8 & \textbf{99.3} & 91.2 \\
    \hline
    \rowcolor{gray!25}
    \textbf{GOAL}               & 95.2 & \textbf{97.0}  & {84.6} & \textbf{100.0} & \textbf{86.9} & 99.1 & {93.8} \\
    \rowcolor{gray!25}
    \textbf{GOAL+CDAN}               & 95.6 & \textbf{97.0}  & \textbf{86.2} & \textbf{100.0} & {86.8} & 99.1 & \textbf{94.1} \\
    \bottomrule
    \toprule
    \textbf{Image-CLEF} & I$\rightarrow$P           & P$\rightarrow$I           & I$\rightarrow$C           & \multicolumn{1}{c}{C$\rightarrow$I}           & \multicolumn{1}{c}{C$\rightarrow$P}           & \multicolumn{1}{c}{P$\rightarrow$C}  & Avg.          \\
    \hline
    Source  \cite{he2016deep}               & 74.8      & 83.9      & 91.5      & 78.0                          & 65.5                          & 91.2                 & 80.7          \\
    DAN      \cite{long2018transferable}             & 74.5      & 82.2      & 92.8      & 86.3                          & 69.2                         & 89.8                 & 82.5          \\
    DANN      \cite{ganin2016domain}            & 75.0      & 86.0      & 96.2      & 87.0                          & 74.3                          & 91.5                 & 85.0          \\
    SAFN \cite{xu2019larger}  & 79.3 &93.3 &96.3&91.7&77.6&95.3&88.9\\
    DWL   \cite{xiao2021dynamic}      & {82.3}     & \textbf{94.8}      &  98.1      & 92.8                          & 77.9                          & \textbf{97.2}        & 90.5          \\
    DSAN \cite{zhu2021deep} & 80.2 & 93.3 & 97.2 & 93.8 & 80.8 & 95.9 & 90.2 \\
    ATM \cite{li2021maximum} & 80.3 & 92.9 & \textbf{98.6} & 93.5 & 77.8 & 96.7 & 90.0 \\
    GDCAN \cite{li2022generalized} & 80.8 & 91.3 & 96.3 & 91.0 & 77.5 & 95.0 & 88.7 \\
    SCDA \cite{li2021semantic} & 80.7 & 92.8 & 96.7 & 93.0 & 81.0 & 95.5 & 90.0 \\
    DMP    \cite{luo2022unsupervised}              & 80.7      & 92.5      & 97.2      & 90.5                          & 77.7                          & 96.2                 & 89.1          \\
    RSDA-MSTN \cite{gu2022unsupervised} & 80.5 & {94.2} & 97.8 & 93.3 & 79.3 & 96.8 & 90.3 \\
    \hline
    \rowcolor{gray!25}
    \textbf{GOAL}                   & 82.2 & 94.1 & 97.3          & {95.6} & {82.3} & 96.4 & {91.3} \\
    \rowcolor{gray!25}
    \textbf{GOAL+CDAN} &  \textbf{82.6} & 94.0 & 97.5 & \textbf{95.7} & \textbf{82.6} & 96.8 & \textbf{91.5} \\         
    \bottomrule
    \end{tabular}
\end{table}

\begin{table*}[t]
    \centering
    \caption{Class-wise classification accuracies (\%) on \textbf{VisDA-2017} (ResNet-101).}
    \label{tab:compare_VisDA}
    \renewcommand{\tabcolsep}{0.6pc} 
    \begin{tabular}{lccccccccccccc}
    \toprule
    \textbf{VisDA-2017} & plane & bcycl & bus & car & horse & knife & mcyle & person & plant & sktbrd & train & truck & Avg. \\
    \hline
    Source \cite{he2016deep}                   & 55.1 &  53.3 &  61.9 &  59.1 &  80.6 &  17.9 &  79.7 &  31.2 &  81.0 &  26.5 &  73.5 &  8.5 &  52.4 \\
    DAN \cite{long2018transferable}                    & 89.2 & 37.2 & 77.7 & 61.8 & 81.7 & 64.3 & 90.6 & 61.4 & 79.9 & 37.7 & 88.1 & 27.4 & 66.4 \\
    DANN \cite{ganin2016domain}                    & 93.5 & 74.3 & 83.4 & 50.7 & 87.2 & 90.2 & 89.9 & 76.1 & 88.1 & 91.4 & 89.7 & 39.8 & 79.5 \\
    MCD \cite{saito2018maximum} & 87.8 & 75.7 & 84.2 & {78.1} & 91.6 & 95.3 & 88.1 & 78.3 & 83.4 & 64.5 & 84.8 & 20.9 & 77.7 \\
    SAFN \cite{xu2019larger}                       & 93.6 & 61.3 & 84.1 & 70.6 & {94.1} & 79.0 & 91.8 & {79.6} & 89.9 & 55.6 & 89.0 & 24.4 & 76.1 \\
    CRST \cite{zou2019confidence} & 88.0 & 79.2 & 61.0 & 60.0 & 87.5 & 81.4 & 86.3 & 78.8 & 85.6 & 86.6 & 73.9 & \textbf{68.8} & 78.1 \\
    BSP+CDAN \cite{chen2019transferability}        & {95.7} & 75.6 & 82.8 & 54.5 & 89.2 & \textbf{96.5} & 91.3 & 72.2 & 88.9 & 88.7 & 88.0 & 43.4 &  80.5 \\
    MDD \cite{zhang2019bridging} & 88.3 & 62.8 & \textbf{85.2} & 69.9 & 91.9 & 95.1 & \textbf{94.4} & 81.2 & \textbf{93.8} & 89.8 & 84.1 & 47.9 & 82.0 \\
    FixMatch \cite{sohn2020fixmatch} & 96.5 & 76.6 & 72.6 & \textbf{84.6} & \textbf{96.3} & 92.6 & 90.5 & 81.8 & 91.9 & 74.6 & 87.3 & 8.6 & 79.5 \\
    MCC \cite{jin2020minimum} & 95.3 & \textbf{85.8} & 77.1 & 68.0 & 93.9 & 92.9 & 84.5 & 79.5 & 93.6 & 93.7 & 85.3 & 53.8 & 83.6 \\
    DSAN \cite{zhu2021deep}                      & 90.9 & 66.9 & 75.7 & 62.4 & 88.9 & 77.0 & {93.7} & 75.1 & 92.8 & 67.6 & {89.1} & 39.4 & 75.1 \\
    GDCAN \cite{li2022generalized} & 90.0 & 64.9 & 81.1 & 53.2 & 89.1 & 92.9 & 79.5 & 70.9 & 73.0 & 66.5 & 82.4 & 24.4 & 72.3 \\
    SCDA \cite{li2021semantic} & \textbf{96.6} & 77.7 & 82.6 & 61.6 & 91.4 & 4.9 & 82.2 & \textbf{83.6} & 92.1 & 70.9 & 84.7 & 25.4 & 71.2 \\
    DMP  \cite{luo2022unsupervised}              & 92.1 & 75.0 & 78.9 & {75.5} & 91.2 & 81.9 & 89.0 & 77.2 & 93.3 & 77.4 & 84.8 & 35.1 & 79.3 \\
    RSDA-DANN \cite{gu2022unsupervised} & 94.5 & 79.0 & 84.1 & 77.6 & 94.4 & 90.3 & 92.3 & 83.2 & 92.2 & 87.3 & 91.3 & 32.2 & 83.2 \\
    \hline
    \rowcolor{gray!25}
    \textbf{GOAL} & {95.4} & 75.7 & 81.5 & 74.6 & 92.9 & {92.6} & 91.7 & 78.4 & 92.2 & {91.6} & 88.0 & {49.4} & {83.7} \\
    \rowcolor{gray!25}
    \textbf{GOAL+CDAN} & {95.5} & 78.0 & 75.2 & 74.1 & 93.7 & \textbf{96.5} & 89.6 & {82.5} & 90.5 & \textbf{94.5} & \textbf{92.3} & {57.6} & \textbf{85.0} \\
    \bottomrule
    \end{tabular}
\end{table*}

\textbf{Office-31}. Table~\ref{tab:compare_31_CLEF} (top) shows results on Office-31. Since there are 31 object categories and some noisy samples on Office-31, the learning of transferability and discriminability is more difficult compared with ImageCLEF. We observe that GOAL still outperforms other baseline methods and achieves the highest accuracy of 93.8\% on average. Specifically, compared with methods based on geometric framework, e.g., SAFN, BSP, DMP and RSDA, GOAL learns rigorously defined geometric properties (i.e., equivalence and orthogonality on subspaces) with theory-driven learning principles (i.e., Theorem~\ref{thm:norm_TB_thm}-\ref{thm:norm_DB_thm}). Note that GOAL also outperforms other norm-based methods which are insufficient to characterize the discriminability and transferability from theoretical aspects, i.e., SAFN, BSP and BMM. Besides, the combination GOAL+CDAN further improves the results to 94.1\%. These results validate that the proposed geometry-oriented principle can learn stronger transferability and discriminability compared with other norm-based principles.

\textbf{Image-CLEF}. Table~\ref{tab:compare_31_CLEF}~(bottom) shows the results on Image-CLEF. Since GOAL learns both transferability and discriminability, there is a significant improvement over other methods which focus on the domain alignment, i.e., DAN, DANN, ETD. Similar to GOAL, DWL considers the interaction between domain alignment and class discrimination, and achieves better results on several tasks. Further, GOAL learns more preferable geometry structure with explicit co-regularization relation, and achieves much higher accuracy on the most difficult C$\rightarrow$P task. Compared with RSDA which aims to mitigate misalignment problem via robust pseudo-label loss, GOAL simultaneously considers the class-level alignment and cluster orthogonality, and ensure better classification performance. Besides, the average accuracies of GOAL-based model are 91.3\% and 91.5\%, which outperform other methods. Since the domain gap on Image-CLEF is smaller than others, these results show the importance of enhancing discriminability of invariant representations, i.e., learning orthogonality in cluster subspaces.

\textbf{VisDA-2017}. Following the standard protocols \cite{peng2017visda}, the transfer task S$\rightarrow$R is conducted on VisDA-2017, and the class-wise accuracies are presented in Table~\ref{tab:compare_VisDA}. Note that VisDA-2017 has the largest sample size compared with others, which implies the transfer task will be harder, and the simultaneous learning of transferability and discriminability will be more crucial. From the results we can observe that GOAL achieves higher average accuracy than other transferability and discriminability enhancement models. Note that compared with other models, GOAL admits the properties and interpretability from the perspective of geometry, which ensures better invariant representations. Besides, compared with DWL which also tries to balance the learning of abilities, GOAL is formulated as a unified norm-based framework, and the co-regularization relation to achieve balance is theoretically proved (i.e., Theorem~\ref{thm:TBDB_CoRegular}). Thus, GOAL and GOAL+CDAN significantly improve the average accuracy to 83.7\% and 85.0\%, respectively. Compared with self-training-based methods, e.g., FixMatch, CRST and MCC, GOAL uses pseudo-labels in a more effective way, which permits better performance, i.e., directly learning the geometric properties of latent representations. The results above also validate the effectiveness of GOAL in learning balanced abilities on large-scale dataset with complex domain shift.

\begin{table*}[t]
    \centering
    \caption{Classification accuracies (\%) on \textbf{DomainNet} (ResNet-101) and \textbf{Office-Home} (ResNet-50).}
    \label{tab:compare_home_DomainNet}
    \renewcommand{\tabcolsep}{0.095pc} 
    \begin{tabular}{cccccccc||cccccccc||cccccccc}
    \toprule
    \textbf{ADDA} \cite{tzeng2017adversarial} & clp & inf & pnt & qdr & rel & skt & Avg. & \textbf{DANN} \cite{ganin2016domain} & clp & inf & pnt & qdr & rel & skt & Avg. & \textbf{MCD} \cite{saito2018maximum} & clp & inf & pnt & qdr & rel & skt & Avg. \\
    \hline
    clp & - & 11.2 & 24.1 & 3.2 & 41.9 & 30.7 & 22.2 & clp & - & 15.5 & 34.8 & 9.5 & 50.8 & 41.4 & 30.4 & clp & - & 14.2 & 26.1 & 1.6 & 45.0 & 33.8 & 24.1 \\
    inf & 19.1 & - & 16.4 & 3.2 & 26.9 & 14.6 & 16.0 & inf & 31.8 & - & 30.2 & 3.8 & 44.8 & 25.7 & 27.3 & inf & 23.6 & - & 21.2 & 1.5 & 36.7 & 18.0 & 20.2 \\
    pnt & 31.2 & 9.5 & - & 8.4 & 39.1 & 25.4 & 22.7 & pnt & 39.6 & 15.1 & - & 5.5 & 54.6 & 35.1 & 30.0 & pnt & 34.4 & 14.8 & - & 1.9 & 50.5 & 28.4 & 26.0 \\
    qdr & 15.7 & 2.6 & 5.4 & - & 9.9 & 11.9 & 9.1 & qdr & 11.8 & 2.0 & 4.4 & - & 9.8 & 8.4 & 7.3 & qdr & 15.0 & 3.0 & 7.0 & - & 11.5 & 10.2 & 9.3  \\
    rel & 39.5 & 14.5 & 29.1 & 12.1 & - & 25.7 & 24.2 & rel & 47.5 & 17.9 & 47.0 & 6.3 & - & 37.3 & 31.2 & rel & 42.6 & 19.6 & 42.6 & 2.2 & - & 29.3 & 27.2 \\
    skt & 35.3 & 8.9 & 25.2 & 14.9 & 37.6 & - & 25.4 & skt & 47.9 & 13.9 & 34.5 & 10.4 & 46.8 & - & 30.7 & skt & 41.2 & 13.7 & 27.6 & 3.8 & 34.8 & - & 24.2 \\
    Avg. & 28.2 & 9.3 & 20.1 & 8.4 & 31.1 & 21.7 & \cellcolor{gray!25} 19.8 & Avg. & 35.7 & 12.9 & 30.2 & 7.1 & 41.4 & 29.6 & \cellcolor{gray!25} 26.1 & Avg. & 31.4 & 13.1 & 24.9 & 2.2 & 35.7 & 23.9 & \cellcolor{gray!25} 21.9 \\
    \bottomrule
    \toprule
    \textbf{BNM} \cite{cui2020towards} & clp & inf & pnt & qdr & rel & skt & Avg. & \textbf{CDAN} \cite{long2018conditional} & clp & inf & pnt & qdr & rel & skt & Avg. & \textbf{MDD} \cite{zhang2019bridging} & clp & inf & pnt & qdr & rel & skt & Avg. \\
    \hline
    clp & - & 12.1 & 33.1 & 6.2 & 50.8 & 40.2 & 28.5 & clp & - & 20.4 & 36.6 & 9.0 & 50.7 & 42.3 & 31.8 & clp & - & 20.5 & 40.7 & 6.2 & 52.5 & 42.1 & 32.4 \\
    inf & 26.6 & - & 28.5 & 2.4 & 38.5 & 18.1 & 22.8 & inf & 27.5 & - & 25.7 & 1.8 & 34.7 & 20.1 & 22.0 & inf & 33.0 & - & 33.8 & 2.6 & 46.2 & 24.5 & 28.0 \\
    pnt & 39.9 & 12.2 & - & 3.4 & 54.5 & 36.2 & 29.2 & pnt & 42.6 & 20.0 & - & 2.5 & 55.6 & 38.5 & 31.8 & pnt & 43.7 & 20.4 & - & 2.8 & 51.2 & 41.7 & 32.0 \\
    qdr & 17.8 & 1.0 & 3.6 & - & 9.2 & 8.3 & 8.0 & qdr & 21.0 & 4.5 & 8.1 & - & 14.3 & 15.7 & 12.7 & qdr & 18.4 & 3.0 & 8.1 & - & 12.9 & 11.8 & 10.8 \\
    rel & 48.6 & 13.2 & 49.7 & 3.6 & - & 33.9 & 29.8 & rel & 51.9 & 23.3 & 50.4 & 5.4 & - & 41.4 & 34.5 & rel & 52.8 & 21.6 & 47.8 & 4.2 & - & 41.2 & 33.5 \\
    skt & 54.9 & 12.8 & 42.3 & 5.4 & 51.3 & - & 33.3 & skt & 50.8 & 20.3 & 43.0 & 2.9 & 50.8 & - & 33.6 & skt & 54.3 & 17.5 & 43.1 & 5.7 & 54.2 & - & 35.0 \\
    Avg. & 37.6 & 10.3 & 31.4 & 4.2 & 40.9 & 27.3 & \cellcolor{gray!25} 25.3 & Avg. & 38.8 & 17.7 & 32.8 & 4.3 & 41.2 & 31.6 & \cellcolor{gray!25} 27.7 & Avg. & 40.4 & 16.6 & 34.7 & 4.3 & 43.4 & 32.3 & \cellcolor{gray!25} 28.6 \\
    \bottomrule
    \toprule
    \textbf{SCDA}\textsuperscript{\textbf{+CDAN}} \cite{li2021semantic} & clp & inf & pnt & qdr & rel & skt & Avg. & \textbf{GOAL} & clp & inf & pnt & qdr & rel & skt & Avg. & \textbf{GOAL}\textsuperscript{\textbf{+CDAN}} & clp & inf & pnt & qdr & rel & skt & Avg. \\
    \hline
    clp & - & 19.5 & 40.4 & 10.3 & 56.7 & 46.0 & 34.6 & clp & - & 19.3 & 37.8 & 14.4 & 53.2 & 43.8 & 33.7 & clp & - & 20.9 & 39.6 & 16.0 & 54.7 & 45.3 & 35.3 \\
    inf & 35.6 & - & 36.7 & 4.5 & 50.3 & 29.9 & 31.4 & inf & 45.5 & - & 35.4 & 7.4 & 51.3 & 33.9 & 34.7 & inf & 47.0 & - & 36.9 & 8.9 & 52.8 & 35.4 & 36.2    \\
    pnt & 45.6 & 20.0 & - & 4.2 & 56.8 & 41.9 & 33.7 & pnt & 50.8 & 22.9 & - & 10.1 & 59.5 & 43.4 & 37.3 & pnt & 52.3 & 24.4 & - & 11.8 & 61.0 & 44.9 & 38.9    \\
    qdr & 28.3 & 4.8 & 11.5 & - & 20.9 & 19.2 & 17.0 & qdr & 24.1 & 2.9 & 7.5 & - & 12.8 & 16.5 & 12.8 & qdr & 25.8 & 4.4 & 9.0 & - & 14.7 & 18.1 & 14.4    \\
    rel & 55.5 & 22.8 & 53.7 & 3.2 & - & 42.1 & 35.5 & rel & 58.9 & 26.6 & 54.2 & 16.2 & - & 45.9 & 40.4 & rel & 60.4 & 28.1 & 55.7 & 18.3 & - & 47.4 & 42.0    \\
    skt & 58.4 & 21.1 & 47.8 & 10.6 & 56.5 & - & 38.9 & skt & 59.2 & 21.8 & 46.9 & 19.8 & 56.8 & - & 40.9 & skt & 60.5 & 23.2 & 48.1 & 21.7 & 58.4 & - & 42.4    \\
    Avg. & 44.7 & 17.6 & 38.0 & 6.6 & 48.2 & 35.8 & \cellcolor{gray!25} 31.8 & Avg. & 47.7 & 18.7 & 36.4 & 13.6 & 46.7 & 36.7 & \cellcolor{gray!25} 33.3 & Avg. & 49.2 & 20.2 & 37.8 & 15.3 & 48.3 & 38.2 & \cellcolor{gray!25} \textbf{34.9}    \\
    \bottomrule
    \end{tabular}
    \\[3pt]
    \renewcommand{\tabcolsep}{0.30pc} 
    \begin{tabular}{lccccccccccccc}
    \toprule
    \textbf{Office-Home} & \multicolumn{1}{l}{Ar$\rightarrow$Cl} & \multicolumn{1}{l}{Ar$\rightarrow$Pr} & \multicolumn{1}{l}{Ar$\rightarrow$Rw} & \multicolumn{1}{l}{Cl$\rightarrow$Ar} & \multicolumn{1}{l}{Cl$\rightarrow$Pr} & \multicolumn{1}{l}{Cl$\rightarrow$Rw} & \multicolumn{1}{l}{Pr$\rightarrow$Ar} & \multicolumn{1}{l}{Pr$\rightarrow$Cl} & \multicolumn{1}{l}{Pr$\rightarrow$Rw} & \multicolumn{1}{l}{Rw$\rightarrow$Ar} & \multicolumn{1}{l}{Rw$\rightarrow$Cl} & \multicolumn{1}{l}{Rw$\rightarrow$Pr} & \multicolumn{1}{l}{Avg.} \\
    \hline
    Source \cite{he2016deep}              & 41.1 & 65.9 & 73.7 & 53.1 & 60.1 & 63.3 & 52.2 & 36.7 & 71.8 & 64.8 & 42.6 & 75.2 & 58.4\\
    DAN   \cite{long2018transferable}               & 45.6 & 67.7 & 73.9 & 57.7 & 63.8 & 66.0 & 54.9 & 40.0 & 74.5 & 66.2 & 49.1 & 77.9 & 61.4   \\
    DANN  \cite{ganin2016domain}               & 53.8 & 62.6 & 74.0 & 55.8 & 67.3 & 67.3 & 55.8 & 55.1 & 77.9 & 71.1 & 60.7 & 81.1 & 65.2    \\
    MCD \cite{saito2018maximum} & 51.7 & 72.2 & 78.2 & 63.7 & 69.5 & 70.8 & 61.5 & 52.8 & 78.0 & {74.5} & 58.4 & 81.8 & 67.8    \\
    SAFN  \cite{xu2019larger}                 & 53.2 & 72.7 & 76.8 & 65.0 & 71.3 & 72.3 & 65.0 & 51.4 & 77.9 & 72.3 & 57.8 & 82.4 & 68.2    \\
    BSP+CDAN \cite{chen2019transferability}  & 54.7 & 67.7 & 76.2 & 61.0 & 69.4 & 70.9 & 60.9 & 55.2 & 80.2 & 73.4 & 60.3 & 81.2 & 67.6   \\
    MDD \cite{zhang2019bridging} & 56.2 & 75.4 & 79.6 & 63.5 & 72.1 & 73.8 & 62.5 & 54.8 & 79.9 & 73.5 & 60.9 & 84.5 & 69.7    \\
    BNM \cite{cui2020towards} & 52.3 & 73.9 & 80.0 & 63.3 & 72.9 & {74.9} & 61.7 & 49.5 & 79.7 & 70.5 & 53.6 & 82.2 & 67.9 \\
    FixMatch \cite{sohn2020fixmatch} & 56.4 & 76.4 & 79.9 & 65.3 & 73.8 & 71.2 & {67.2} & {56.4} & 80.6 & {74.9} & \textbf{63.5} & 84.3 & 70.8 \\
    MCC \cite{jin2020minimum} & {58.4} & \textbf{79.6} & 83.0 & {67.5} & {77.0} & \textbf{78.5} & {66.6} & 54.8 & 81.8 & {74.4} & {61.4} & {85.6} & {72.4} \\
    DSAN \cite{zhu2021deep} & {54.4} & 70.8 & 75.4 & 60.4 & 67.8 & 68.0 & 62.6 & {55.9} & 78.5 & {73.8} & {60.6} & 83.1 & 67.6 \\
    ATM \cite{li2021maximum} & 52.4 & 72.6 & 78.0 & 61.1 & 72.0 & 72.6 & 59.5 & 52.0 & 79.1 & 73.3 & 58.9 & 83.4 & 67.9 \\
    GDCAN \cite{li2022generalized} & 57.3 & 75.7 & 83.1 & 68.6 & 73.2 & 77.3 & 66.7 & 56.4 & 82.2 & 74.1 & 60.7 & 83.0 & 71.5 \\
    SCDA+CDAN \cite{li2021semantic} & 57.1 & 75.9 & 79.9 & 66.2 & 76.7 & 75.2 & 65.3 & 55.6 & 81.9 & 74.7 & 62.6 & 84.5 & 71.3 \\
    DMP \cite{luo2022unsupervised}   & 52.3                      & 73.0                      & 77.3                      & {64.3}             & 72.0                      & 71.8                      & 63.6                      & 52.7                      & 78.5                      & 72.0                      & 57.7                      & 81.6                      & 68.1                     \\
    RSDA-MSTN \cite{gu2022unsupervised} & \textbf{59.6} & 79.2 & 81.1 & \textbf{68.7} & \textbf{77.7} & 77.7 & \textbf{67.8} & \textbf{61.0} & 82.2 & \textbf{75.3} & 60.8 & \textbf{85.9} & \textbf{73.1} \\
    \hline
    \rowcolor{gray!25}
    \textbf{GOAL}                & 52.1                      & {75.6}             & 77.4                      & 61.6                      & {74.6}             & 73.7             & {65.0}             & 53.1             & {80.8}             & 69.5                      & 55.9                      & {83.6}             & {68.6}            \\
    \rowcolor{gray!25}
    \textbf{GOAL+CDAN}                & 52.6                      & {75.7}             & 79.2                      & 62.4                      & {74.4}             & {75.0}             & {66.4}             & {56.0}             & \textbf{82.3}             & 70.6                      & 57.8                      & {84.4}             & {69.7}            \\
    \bottomrule
    \end{tabular}
\end{table*}

\textbf{DomainNet}. Table~\ref{tab:compare_home_DomainNet} shows the results on challenging DomainNet dataset. Since DomainNet consists of much more classes and samples, several UDA methods, e.g., DANN and MCD, may fail to learn transferability and lead to performance degradation compared with source model. Though BNM tries to learn diverse predictions, it is indeed insufficient to ensure discriminability when there are significantly more classes. To deal with the challenges in DomainNet, learning the cluster structure of considerable classes is necessary. Therefore, the semantic information learning methods, e.g., CDAN and SCDA+CDAN, improve the accuracies of source model by 1\%$\sim$5\%. Compared with these methods, GOAL simultaneously ensures the transferability and discriminability, which are crucial for learning the complex data structure in DomainNet. Then GOAL and GOAL+CDAN achieve significant performance improvement and the mean accuracies are 33.3\% and 34.9\%. These results validate the effectiveness of proposed geometric principles in challenging learning scenario.

\begin{figure*}[t]
    \centering
    \subfigure[Image-CLEF I$\rightarrow$P \label{fig:3D_clefIP}]{\includegraphics[width=0.245\linewidth,height=90pt,trim=120 60 100 70,clip]{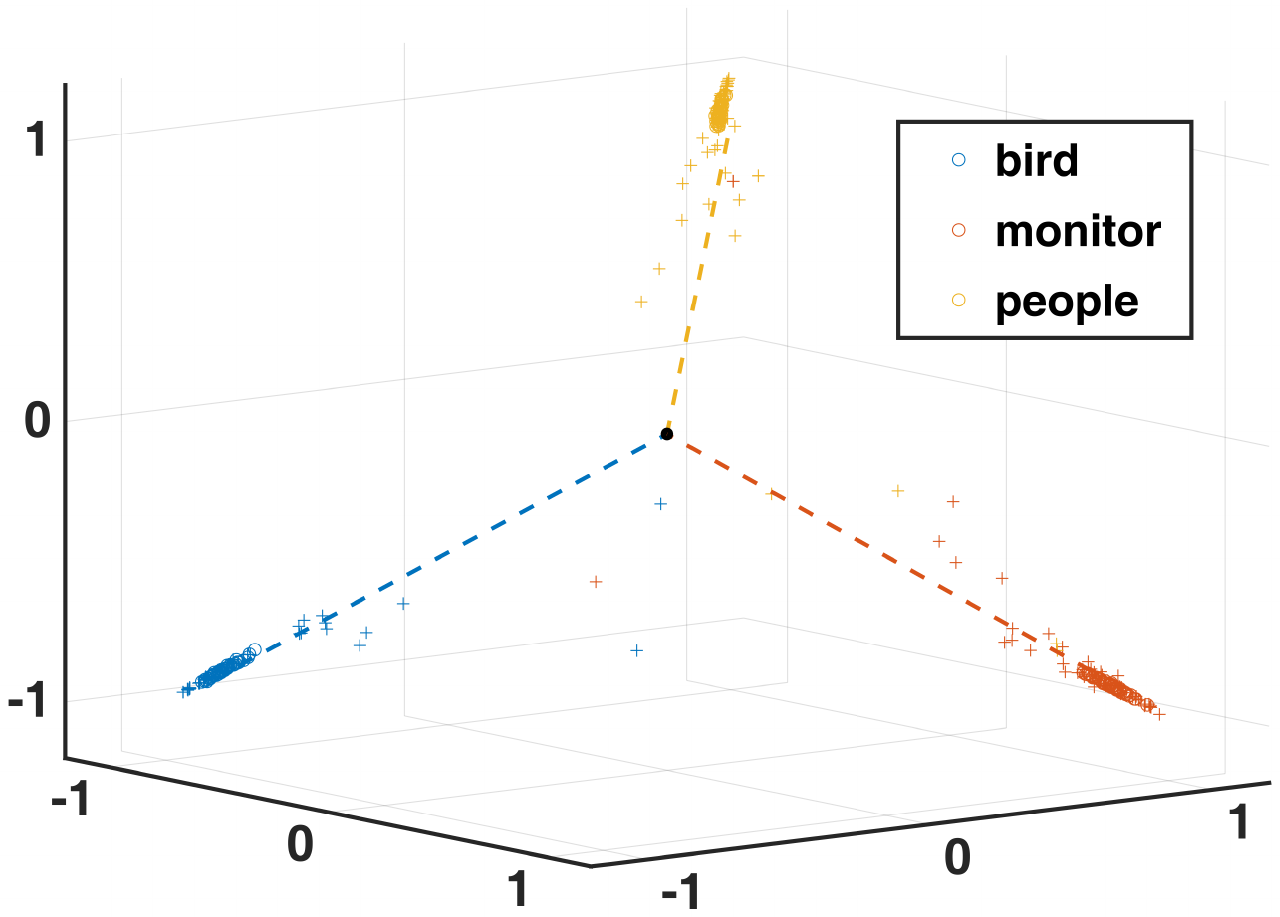}}
    \hfill
    \subfigure[Office-31 A$\rightarrow$W \label{fig:3D_31AW}]{\includegraphics[width=0.245\linewidth,height=90pt,trim=120 60 100 70,clip]{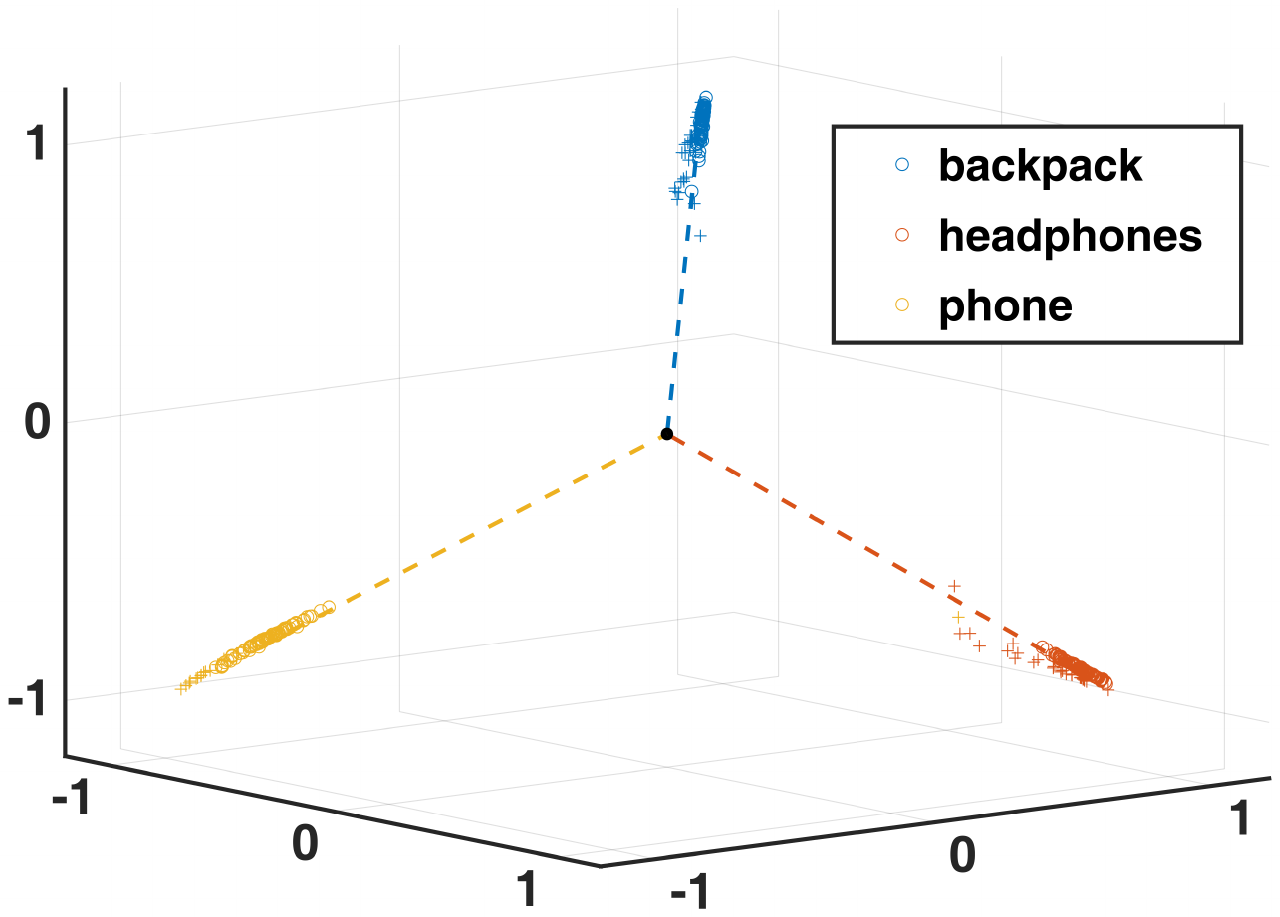}}
    \hfill
    \subfigure[Image-CLEF I$\rightarrow$P \label{fig:hyper_clef}]{\includegraphics[width=0.245\linewidth,height=90pt,trim=150 100 90 110,clip]{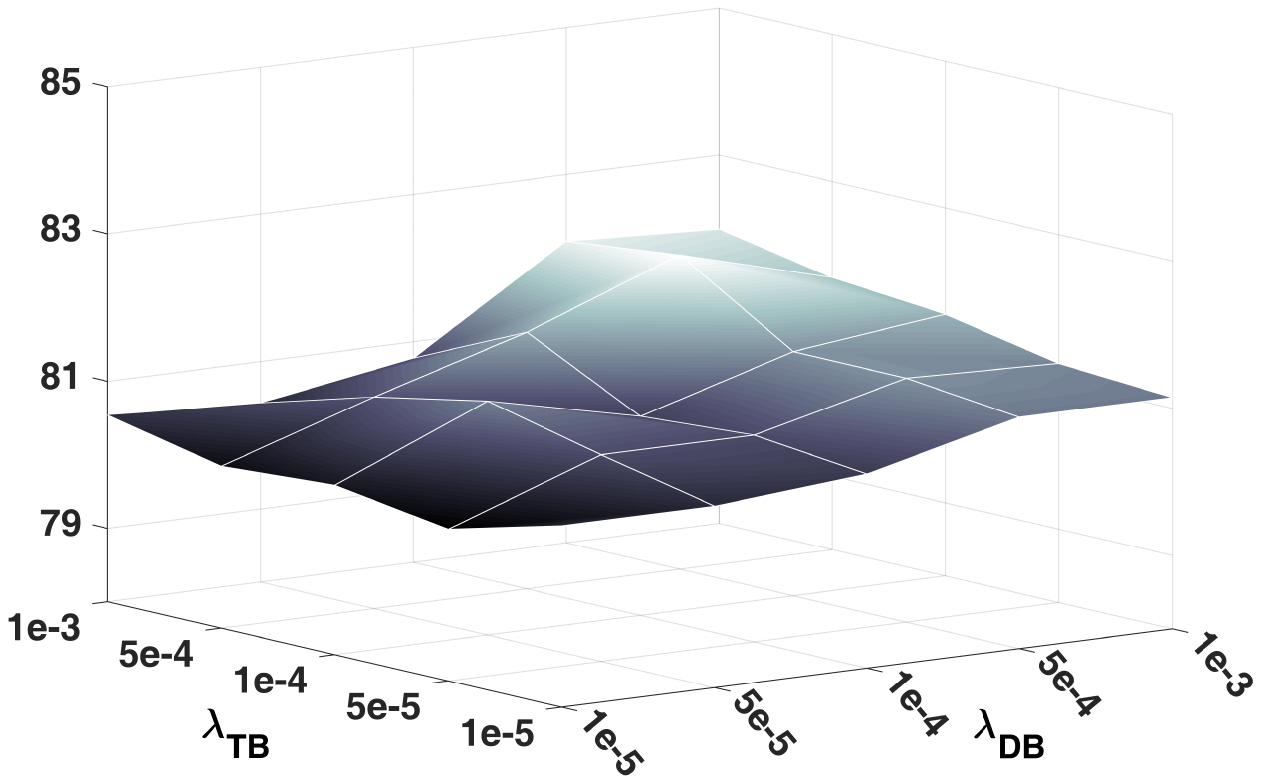}}
    \hfill
    \subfigure[Office-31 A$\rightarrow$W \label{fig:hyper_31}]{\includegraphics[width=0.245\linewidth,height=90pt,trim=150 100 90 110,clip]{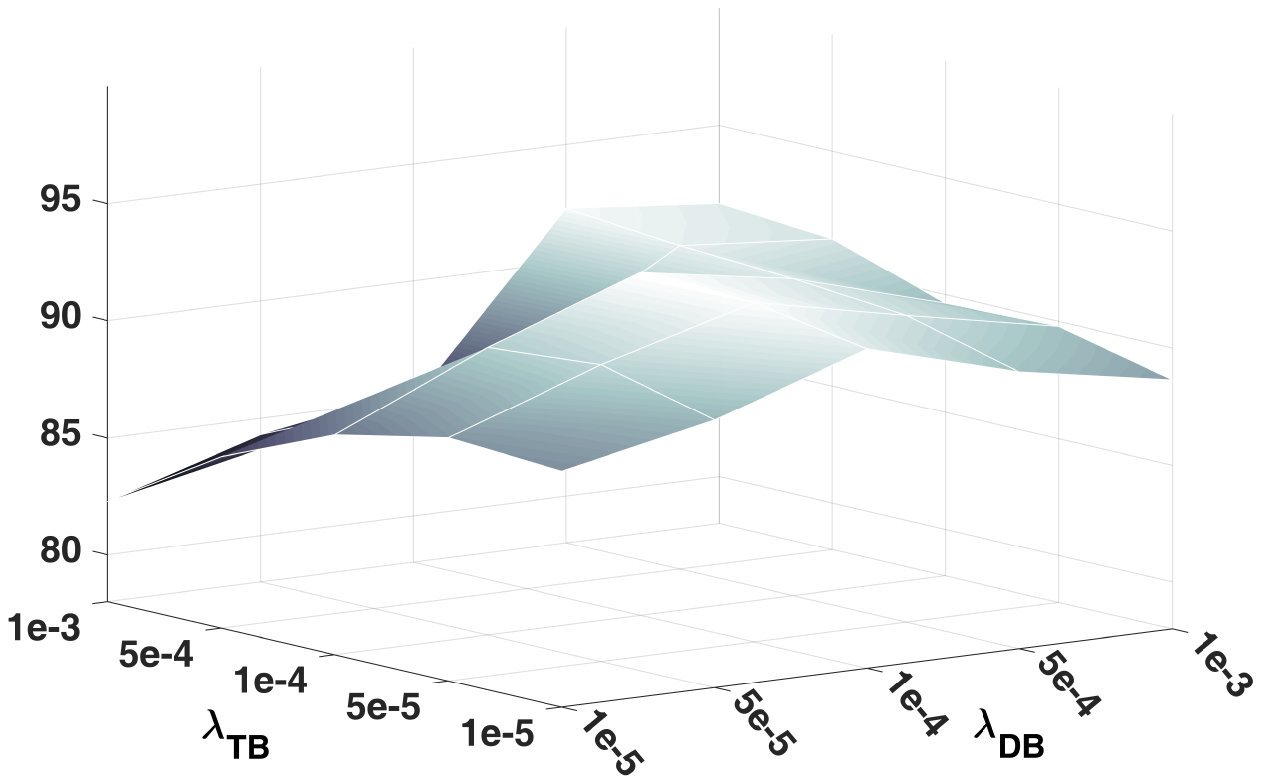}}
    \\
    \caption{\textbf{(a)}-\textbf{(b)}: Visualization of the class orthogonality and the class-level domain equivalence. \textbf{(c)}-\textbf{(d)}: Visualization of the co-regularization relation by varying the regularization parameters.}
    \label{fig:Geometry&Hyper}
\end{figure*}

\textbf{Office-Home}. Table~\ref{tab:compare_home_DomainNet} shows the results on Office-Home dataset. Among the four datasets for evaluation, Office-Home has more categories and large domain divergence, which increases the difficulty of enhancing discriminability. Compared with Source model without adaptation, transferability enhancement models, e.g., DAN, DANN, SAFN and DSAN, achieve higher accuracy by learning invariant representations. Besides, the models considering discriminability, e.g., ETD, DMP and BNM, also achieve significant accuracy improvement by encoding intrinsic structure information into representations. Further, the average classification accuracy of GOAL and GOAL+CDAN are 68.6\% and 69.7\%, which are higher than other baseline methods. Especially, compared with other norm-based learning models, i.e., SAFN, BSP and BNM, the improvements of GOAL are about 2\%$\sim$3\%. It demonstrates that GOAL can improve accuracies by learning enhanced discriminability with nearly orthogonal clusters. Besides, the GOAL has lower accuracy than other advanced methods, e.g., GDCAN, SCDA and RSDA, where the main reason could be that the initialized pseudo-labels are less reliable in Office-Home datasets. Thus, the geometric structure learning could be insufficient, which leads to less performance gain. Note that though GOAL is less significant on Office-Home, it consistently outperforms these baseline methods on other datasets.

Overall, the comparison experiments validate that the theoretical results and proposed algorithm on geometric structure learning are generally effective on different UDA datasets. Moreover, for GOAL with explicit learning principles, the simultaneously achieved transferability and discriminability actually ensure competitive performance compared with SOTA methods. Though the challenging datasets (e.g., VisDA-2017, Office-Home and DomainNet) with low-quality pseudo-labels have impacts on the improvement of applying GOAL, the model can be significantly enhanced by introducing advanced adaptation modules into GOAL. Specifically, the combination of GOAL and CDAN significantly enhances the target classification accuracies by about 1.5\%$\sim$5\%, which demonstrates that the proposed geometric principles can be successfully applied to existing SOTA model and better adaptation performance will be ensured.

\subsection{Empirical Analysis}
In this section, we conduct experiments to validate the theoretical results and methodological analysis of GOAL.

\textbf{Geometric Properties}. To validate the geometric principles $\mathcal{L}_{\text{TB}}$ and $\mathcal{L}_{\text{DB}}$ for learning domain equivalence and class orthogonality, a toy example is presented. We randomly select 3 classes from Image-CLEF I$\rightarrow$P and Office-31 A$\rightarrow$W, and optimize GOAL on 3-dimensional space. Note that from the perspective of subspace geometry, we focus on the angle (i.e., the similarity) between samples. As shown in Figure~\ref{fig:3D_clefIP}, the samples from the same class are gathered together on the same axis (1-dimensional subspaces) while the clusters (axes) are nearly orthogonal. This result demonstrates that $\mathcal{L}_{\text{DB}}$ actually encourages the model to learn maximum discriminability with class orthogonality. Besides, note that the source and target samples are distributed along the same axis in Figure~\ref{fig:3D_31AW}, which verifies that the class-wise domain equivalence can be effectively learned via $\mathcal{L}_{\text{TB}}$. These results validate the norm-based principles indeed ensure favorable geometry structures.

\begin{figure*}[t]
    \centering
    \subfigure[Image-CLEF I$\rightarrow$P \label{fig:hyperTheory_clef}]{\includegraphics[width=0.245\linewidth,trim=20 20 20 10,clip]{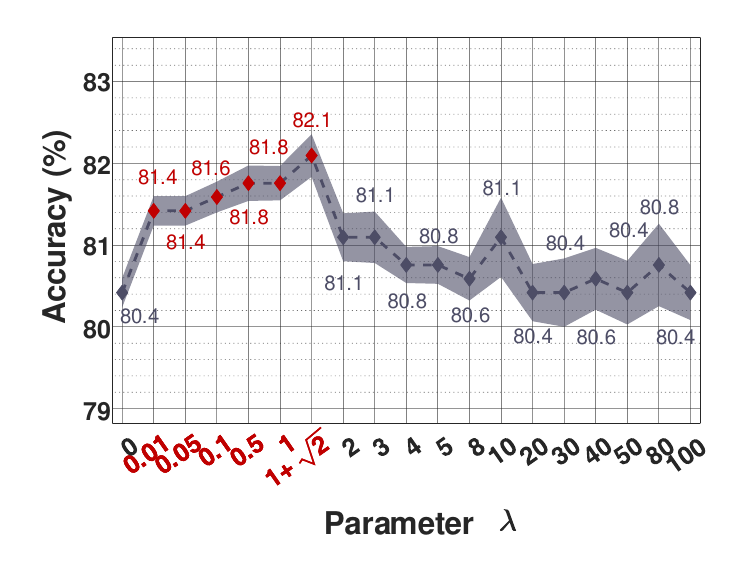}}
    \hfill
    \subfigure[Office-31 A$\rightarrow$W \label{fig:hyperTheory_31}]{\includegraphics[width=0.245\linewidth,trim=20 20 20 10,clip]{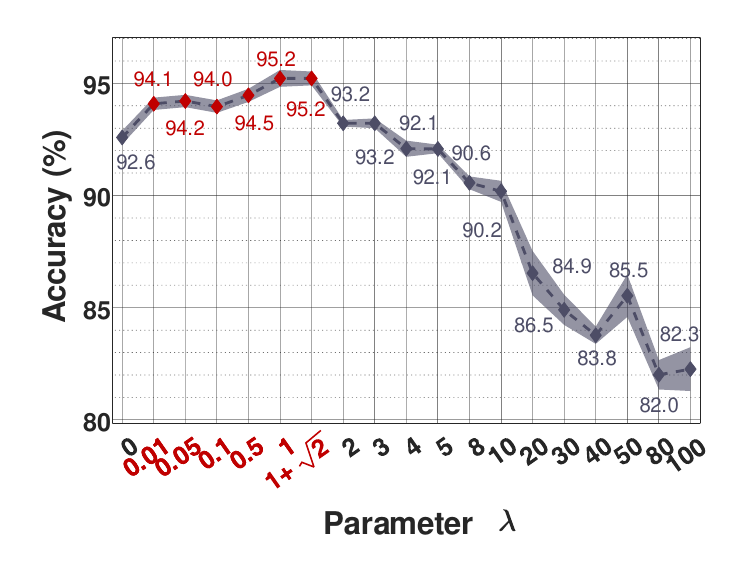}}
    \hfill
    \subfigure[Office-Home Ar$\rightarrow$Cl \label{fig:hyperTheory_home}]{\includegraphics[width=0.245\linewidth,trim=20 20 20 10,clip]{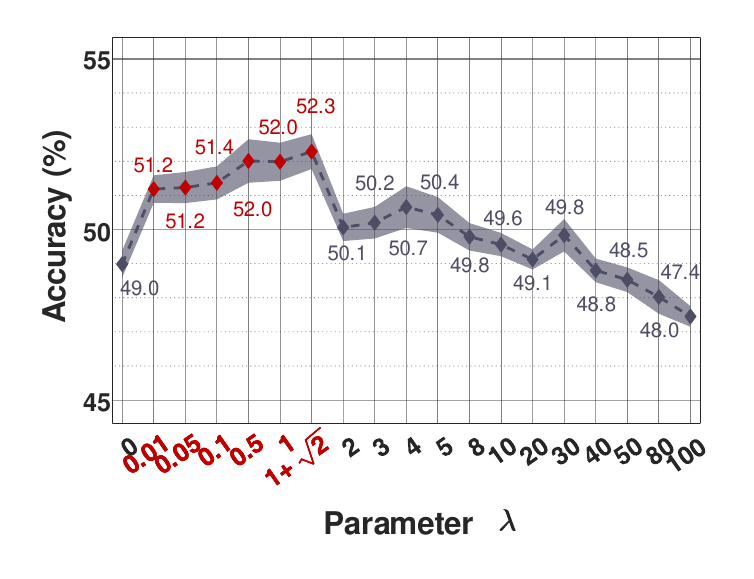}}
    \hfill
    \subfigure[VisDA-2017 S$\rightarrow$R \label{fig:hyperTheory_visda}]{\includegraphics[width=0.245\linewidth,trim=20 20 20 10,clip]{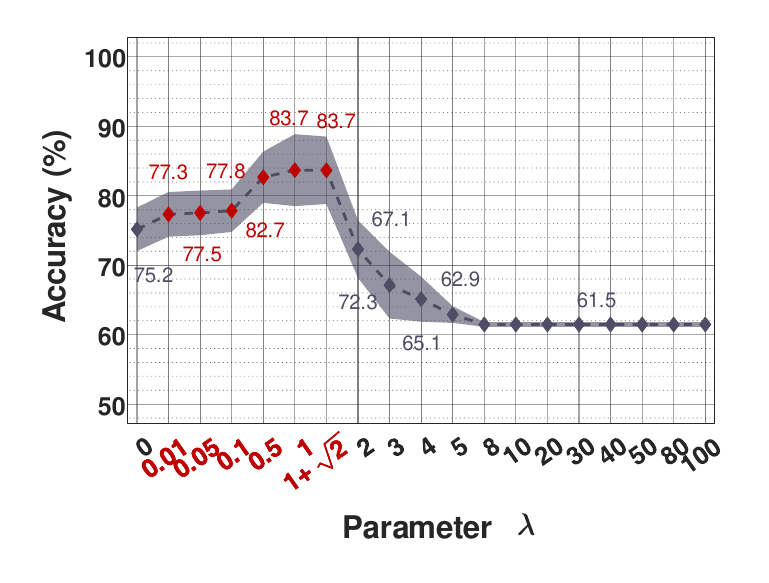}}
    \\
    \subfigure[Office-Home Ar$\rightarrow$Cl \label{fig:pesudo_Tacc_home_A2C}]{\includegraphics[width=0.245\linewidth,height=85pt,trim=50 230 65 255,clip]{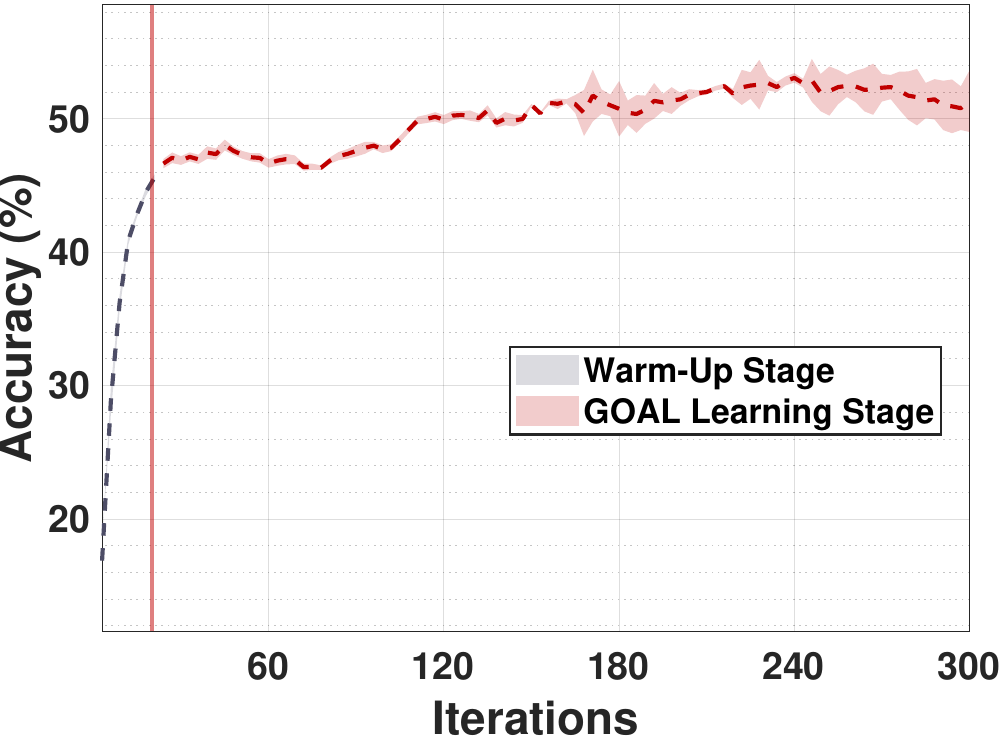}}
    \hfill
    \subfigure[Office-Home Pr$\rightarrow$Cl \label{fig:pesudo_Tacc_home_P2C}]{\includegraphics[width=0.245\linewidth,,height=85pt,trim=50 230 65 255,clip]{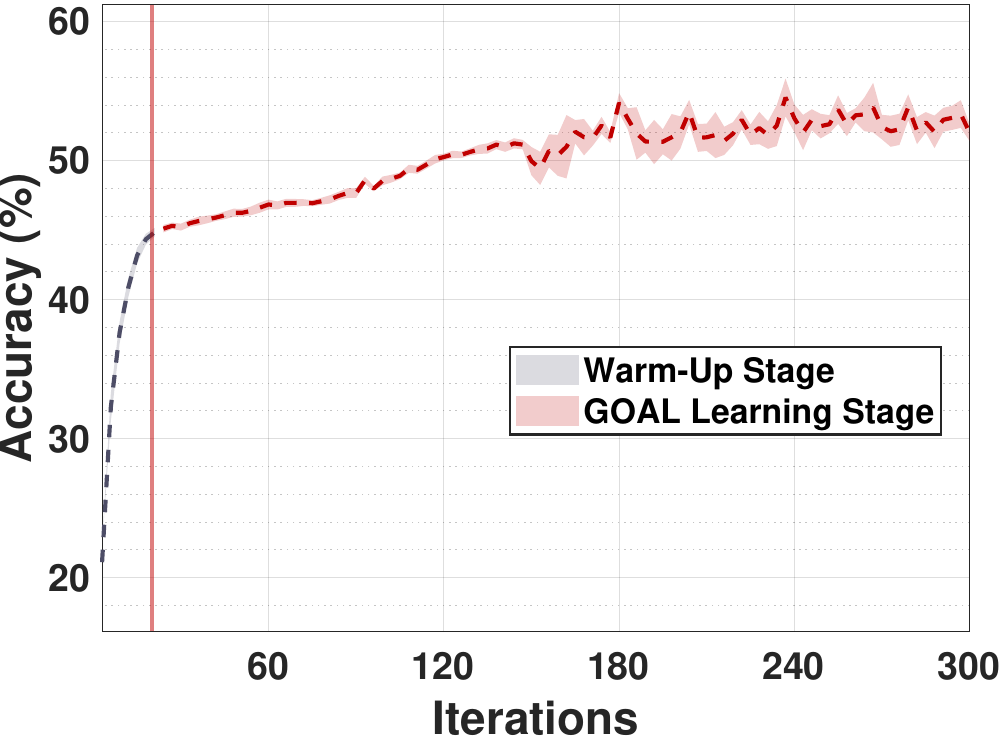}}
    \hfill
    \subfigure[Office-Home Ar$\rightarrow$Cl \label{fig:pesudo_Selection_home_A2C}]{\includegraphics[width=0.245\linewidth,height=88pt,trim=35 230 30 245,clip]{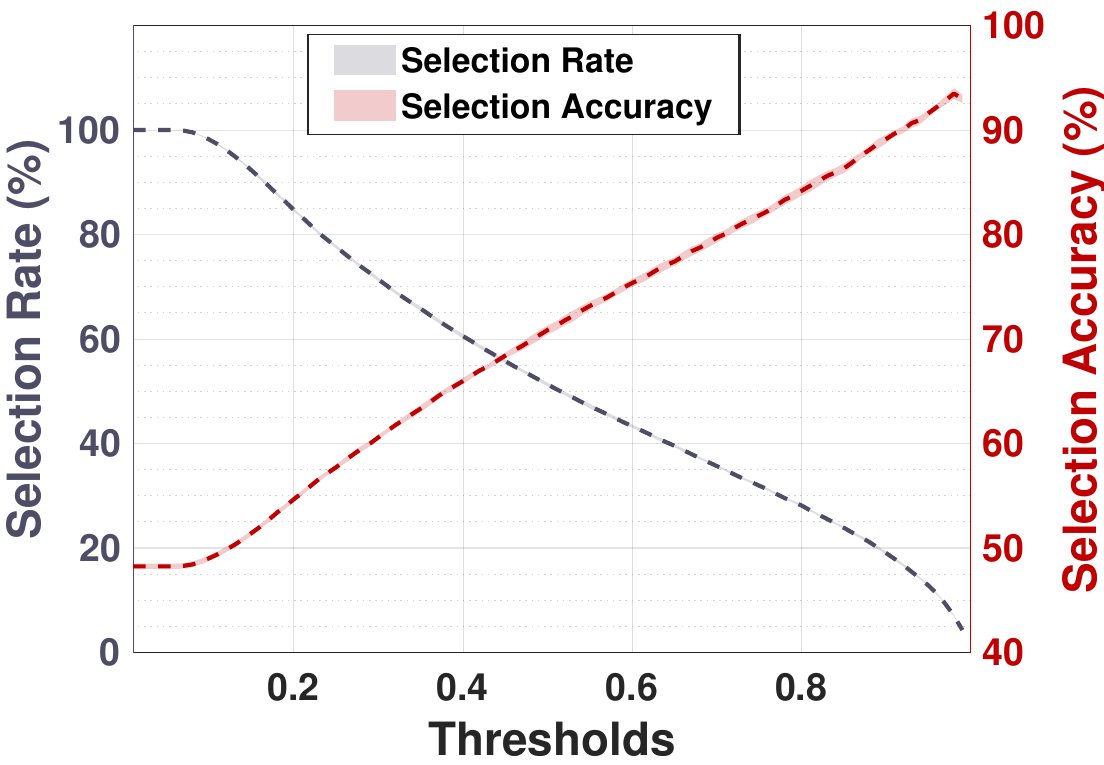}}
    \hfill
    \subfigure[Office-Home Pr$\rightarrow$Cl \label{fig:pesudo_Selection_home_P2C}]{\includegraphics[width=0.245\linewidth,height=88pt,trim=35 230 30 245,clip]{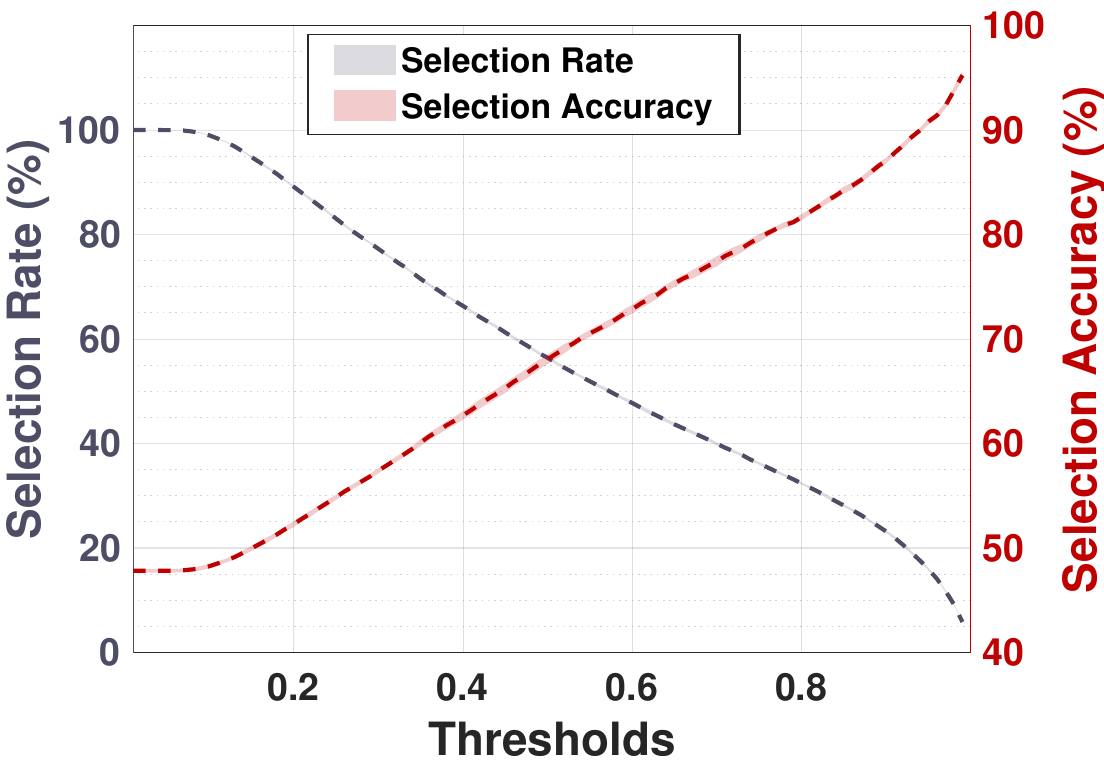}}
    \\
    \caption{\textbf{(a)}-\textbf{(d)}: Visualization of co-regularization by varying $\lambda = \lambda_{\text{TB}}/\lambda_{\text{DB}}$. The better performance is actually achieved when $\lambda\in (0,1+\sqrt{2}]$ (red points). \textbf{(e)}-\textbf{(h)}: Pseudo-label analysis by visualizing the accuracies of pseudo-labels on target domain (i.e., {(e)}-{(f)}) and the selection rates/accuracies under different confidence thresholds (i.e., {(g)}-{(h)}).}
    \label{fig:hyper_lambda}
\end{figure*}

\textbf{Co-regularization}. We validate the co-regularization relation between transferability and discriminability, as shown in Theorem~\ref{thm:TBDB_CoRegular}, by evaluating GOAL model under different hyper-parameters $\lambda_{\text{TB}}$ and $\lambda_{\text{DB}}$. The experiment results on are shown in Figure~\ref{fig:hyper_clef}-\ref{fig:hyper_31}. When $\lambda_{\text{TB}} > \lambda_{\text{DB}}$, it can be seen that the model is sensitive to the hyper-parameters and the accuracies decrease rapidly, which verifies the case \textbf{(\romannumeral2)} in Theorem~\ref{thm:TBDB_CoRegular}. In contrast, when $\lambda_{\text{TB}}\leq \lambda_{\text{DB}}$ (i.e., $\lambda \leq 1$), the model generally robust to the change of hyperparameters. Besides, the highest accuracies are usually achieved when the parameters are balanced, i.e., $\lambda_{\text{TB}} \approx \lambda_{\text{DB}}$. These results validate our theoretical results in Theorem~\ref{thm:TBDB_CoRegular}\textbf{(\romannumeral3)} which suggests that $\lambda_{\text{TB}}$ should be smaller to achieve better geometry structure. Further, the co-regularization is explicitly visualized in Figure~\ref{fig:hyper_lambda} by varying $\lambda$ on Office-Home, Office-31, ImageCLEF and VisDA-2017 datasets. It can be observed that the model actually achieves higher accuracies when $\lambda\in (0,1+\sqrt{2}]$ (i.e., red points). This observation is consistent with our main result in Theorem~\ref{thm:TBDB_CoRegular}. Especially, when $\lambda=0 $ or $\lambda\in (1+\sqrt{2},\infty]$, there will be a significant performance degradation. It implies that model may suffer from the trivial solution as shown in Figure~\ref{fig:co_regularization_illustration}, which leads to insufficiently learned geometric abilities. Overall, results above demonstrate that \textit{the transferability and discriminability are possible to simultaneously achieve}, and the proper parameter range derived in Theorem~\ref{thm:TBDB_CoRegular} is valid and effective.

\begin{table}[t]
    \centering
    \caption{Quantitative analysis of the learned representations.}
    \label{tab:quantitative_analysis}
    \renewcommand{\tabcolsep}{0.6pc}
    \begin{tabular}{c|ccc|cc}
    \toprule
    \multirow{2}*{\textbf{Measures}} & \multicolumn{3}{c|}{LDA Values} & \multicolumn{2}{c}{Cosine Values} \\
     & Inter. & Intra. & Discri. & P. Angle & C. Angle  \\
    \midrule
    w/o $\mathcal{L}_{\text{GO}}$ & 0.086 & 0.912 & 0.095 & 0.670 & 0.707 \\
    w/ $\mathcal{L}_{\text{GO}}$ & \textbf{0.537} & \textbf{0.450} & \textbf{1.194} & \textbf{0.994} & \textbf{0.909} \\
    \bottomrule
    \end{tabular}
\end{table}

\begin{figure}[t]
    \centering
    \includegraphics[width=0.93\linewidth]{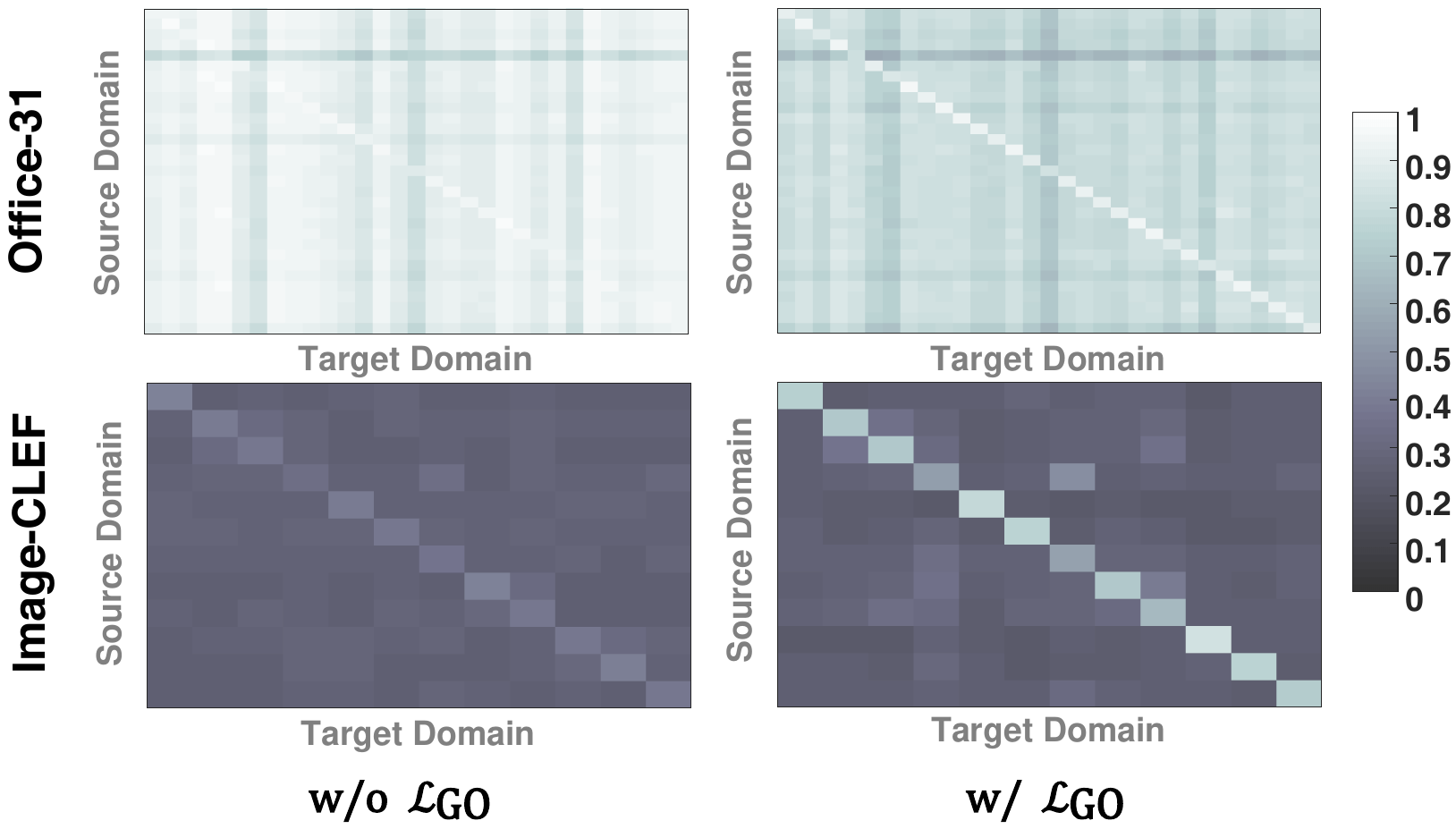} 
    \caption{Visualization of cosine values of pair-wise principal angles. Larger angle values imply more correlated domain subspaces.}
    \label{fig:principal_angle_matrix}
\end{figure}

\textbf{Measures on Vector Spaces}. To quantitatively analyze the transferability and discriminability of learned representations, several measures in subspace analysis are employed. Specifically, we employ similar criteria adopted in recent empirical study \cite{chen2019transferability}. 1) Based on LDA, the inter-class scatter (Intra.), intra-class scatter (Inter.) and discriminant value (Discri.) on the target representations are used to measure the discriminability; 2) The cosine values of principal angle (P. Angle) \cite{bjorck1973numerical} and the class-wise P. Angle (C. Angle) between domains are computed for transferability. The experiment results on Image-CLEF dataset are presented in Table~\ref{tab:quantitative_analysis}. Note that only Intra. values are supposed to be small and others should be large. We see that the model without $\mathcal{L}_{\text{GO}}$ (w/o $\mathcal{L}_{\text{GO}}$) have a large Intra. value and smaller Discri. value, which implies the risk objectives cannot ensure the desirable properties of hidden representations. With the geometry-oriented principle $\mathcal{L}_{\text{GO}}$, the model explicitly learns discriminative and transferable representations, and the LDA values on the target domain are significantly improved. Besides, the cosine values are close to 1, which implies the bases between domains are nearly equivalent. These results further verify the visualization in Figure~\ref{fig:3D_clefIP}-\ref{fig:3D_31AW}, which demonstrates that the proposed principle generally ensures the geometric properties of representations.

\textbf{Visualization of Principal Angles}. The transferability and discriminability are also visualized as heatmaps, which provide more intuitive explanations on geometric relations between clusters. We compute the pair-wise principal angles between $\mathbf{Z}_i^s$ and $\mathbf{Z}_j^t$ (denote as ${\theta}_{ij}$) and visualize $\cos{\theta}$ on Office-31 and Image-CLEF. As shown in Figure~\ref{fig:principal_angle_matrix}, compared with GOAL w/o $\mathcal{L}_{\text{GO}}$, the model with $\mathcal{L}_{\text{GO}}$ ensures more explicit diagonal structure in the heatmap. Specifically, for the model w/ $\mathcal{L}_{\text{GO}}$, the diagonal values are close to 1, which implies the maximum transferability is nearly achieved. Though Office-31 consists of more clusters than Image-CLEF, which increases the difficulty of representation learning, $\mathcal{L}_{\text{GO}}$ still ensures the diversity and identifiability between clusters. Besides, most of the off-diagonal values are close to 0 on Image-CLEF. It demonstrates that the inter-class subspaces are nearly orthogonal, and the clusters are discriminative. In conclusion, the small diagonal values and large off-diagonal values validated the transferability and discriminability learned by $\mathcal{L}_{\text{GO}}$.

\textbf{Effectiveness of Pseudo-Labels}. To study the impacts of pseudo-label strategy, we conduct experiments on Office-Home to observe the pseudo-labeling accuracies, selection rates and selection accuracies. As shown in Figure~\ref{fig:pesudo_Tacc_home_A2C}-\ref{fig:pesudo_Tacc_home_P2C}, compared with the model without warm-up (i.e., source-only training), warm-up training improves the accuracies of pseudo-labels by about 10\%$\sim$14\% and achieves the accuracy of 45\% in average. This result demonstrates that warm-up training can achieve preliminary adaptation and improve the accuracy of source model significantly. Besides, results in Figure~\ref{fig:pesudo_Selection_home_A2C}-\ref{fig:pesudo_Selection_home_P2C} show that the accuracies of selected pseudo-labels can be improved with higher confidence threshold, e.g., the accuracies are about 70\% and 85\% with $\tau=0.5$ and $\tau=0.8$, respectively. Meanwhile, there are about 55\% pseudo-labels are selected when $\tau=0.5$, which implies the confidence threshold can ensure reasonable sample-size and precision of selected pseudo-labels. In conclusion, warm-up training can actually improve the quality of pseudo-labels compared with source model, and confidence selection strategy can further reduce the risk of uncertain pseudo-labels.

\begin{figure}[t]
    \centering
    \subfigure[Office-31 D$\rightarrow$W \label{fig:SVD_AllCls_31DW}]{\includegraphics[width=0.495\linewidth,trim=65 230 65 250,clip]{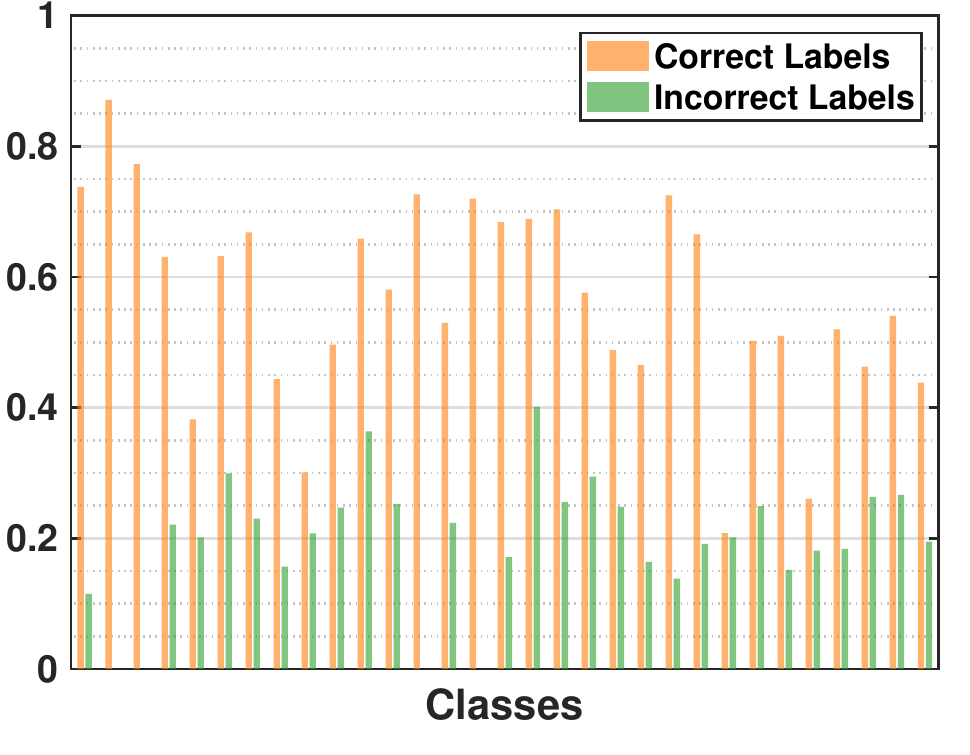}}
    \hfill
    \subfigure[VisDA-2017 S$\rightarrow$R \label{fig:SVD_AllCls_VisDA}]{\includegraphics[width=0.495\linewidth,trim=65 230 65 250,clip]{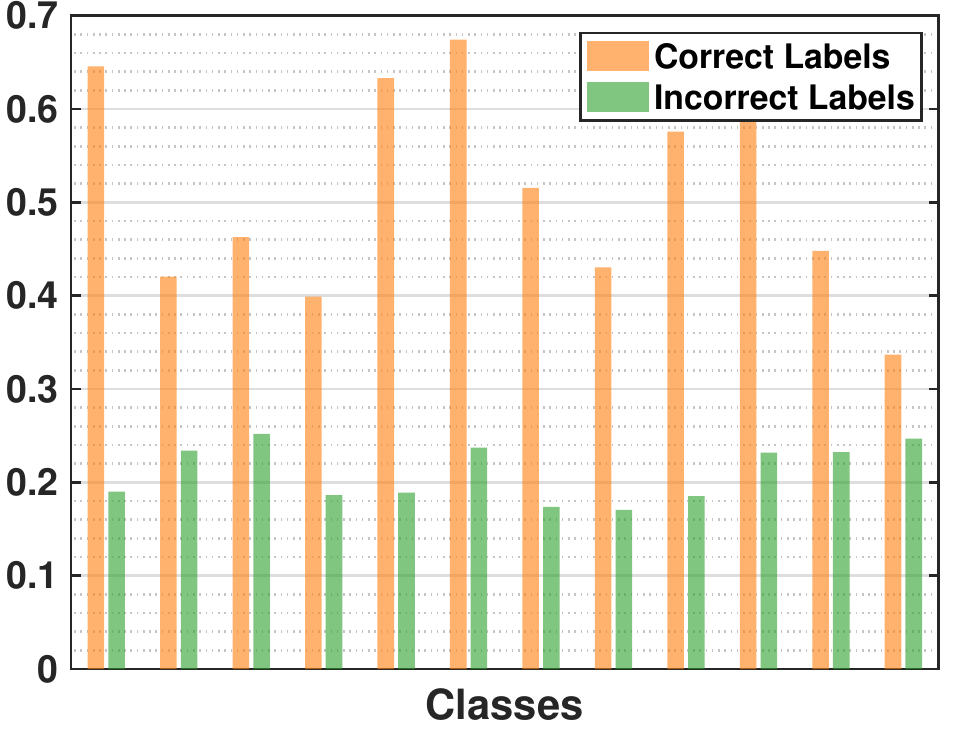}}
    \\
    \caption{Mean values of cosine similarities between correctly/incorrectly labeled samples and dominant singular vectors, where the class matrices $\mathbf{Z}_i$ are analyzed separately.}
    \label{fig:SVD_pseudo_label}
\end{figure}

\begin{table*}[t]
    \centering
    \caption{Model evaluations on challenging adaptation scenario with low-quality pseudo-labels, where warm-up is abbreviated as WU. Rows and columns imply the source domains and target domains, respectively.}
    \label{tab:PseudoLabel_DomainNet}
    \renewcommand{\tabcolsep}{0.13pc} 
    \begin{tabular}{cccccccc||cccccccc||cccccccc}
    \toprule
    \textbf{Source} \cite{he2016deep} & clp & inf & pnt & qdr & rel & skt & Avg. & \textbf{GOAL}\textsuperscript{\textbf{w/o WU}} & clp & inf & pnt & qdr & rel & skt & Avg. & \textbf{GOAL}\textsuperscript{\textbf{w/ WU}} & clp & inf & pnt & qdr & rel & skt & Avg. \\
    \hline
    clp & - & 19.3 & 37.5 & 11.1 & 52.2 & 41.0 & 32.2 & clp & - & 18.8 & 36.3 & 9.1 & 55.7 & 34.3 & 30.9 & clp & - & 19.3 & 37.8 & 14.4 & 53.2 & 43.8 & 33.7 \\
    inf & 30.2 & - & 31.2 & 3.6 & 44.0 & 27.9 & 27.4 & inf & 31.0 & - & 34.8 & 5.9 & 52.8 & 24.1 & 29.7 & inf & 45.5 & - & 35.4 & 7.4 & 51.3 & 33.9 & 34.7 \\
    pnt & 39.6 & 18.7 & - & 4.9 & 54.5 & 36.3 & 30.8 & pnt & 37.9 & 19.0 & - & 6.1 & 61.5 & 32.0 & 31.3 & pnt & 50.8 & 22.9 & - & 10.1 & 59.5 & 43.4 & 37.3  \\
    qdr & 7.0 & 0.9 & 1.4 & - & 4.1 & 8.3 & 4.3 & qdr & 11.0 & 5.3 & 7.1 & - & 10.2 & 9.9 & 8.7 & qdr & 24.1 & 2.9 & 7.5 & - & 12.8 & 16.5 & 12.8 \\
    rel & 48.4 & 22.2 & 49.4 & 6.4 & - & 38.8 & 33.0 & rel & 43.5 & 20.9 & 48.6 & 6.9 & - & 31.9 & 30.3 & rel & 58.9 & 26.6 & 54.2 & 16.2 & - & 45.9 & 40.4   \\
    skt & 46.9 & 15.4 & 37.0 & 10.9 & 47.0 & - & 31.4 & skt & 43.5 & 17.6 & 39.2 & 9.8 & 52.8 & - & 32.6 & skt & 59.2 & 21.8 & 46.9 & 19.8 & 56.8 & - & 40.9  \\
    Avg. & 34.4 & 15.3 & 31.3 & 7.4 & 40.4 & 30.5 & \cellcolor{gray!25} 26.6 & Avg. & 33.4 & 16.3 & 33.2 & 7.6 & 46.6 & 26.4 & \cellcolor{gray!25} 27.3 & Avg. & 47.7 & 18.7 & 36.4 & 13.6 & 46.7 & 36.7 & \cellcolor{gray!25} 33.3 \\
    \bottomrule
    \end{tabular}
\end{table*}

\textbf{Pseudo-Labels and Dominant Basis}. To validate the methodological analysis of pseudo-labels in Section~\ref{subsec:GOAL_model}, the experiment is conducted on Office-31 and VisDA-2017. Generally, the mean values of cosine similarities between correctly/incorrectly labeled samples and dominant singular vectors are computed, which will show weather nuclear norm is decided by the correctly labeled samples or not. Specifically, for a given class matrix, e.g., $\mathbf{Z}_i$, let $(\sigma^{(i)}_j,\mathbf{u}^{(i)}_j)$ be its singular values and corresponding left singular vector. We first compute the mean similarity between singular vector $\mathbf{u}^{(i)}_j$ and correctly/incorrectly labeled samples, and denote it as $s^{(i)}_j$. Then for all singular pairs $(\sigma^{(i)}_j,\mathbf{u}^{(i)}_j)$ of $\mathbf{Z}_i$, the weighted sum of cosine similarities $s^{(i)}_j$ are computed based on their contributions, i.e., $\sum_j  \frac{\sigma^{(i)}_j}{\sum_l \sigma^{(i)}_l} s^{(i)}_j$.

From the results in Figure~\ref{fig:SVD_pseudo_label}, we can conclude two important conclusions. 1) Compared with incorrectly labeled samples, the correctly labeled samples have significantly larger similarities with dominant singular vectors. Thus, when considering pseudo-labels for geometric learning, i.e., applying pseudo-labels to construct the class matrices $\mathbf{Z}_i$, the nuclear norm-based principles can be still effective and reliable. 2) As we weighted the similarities with the contributions of singular vectors, i.e., singular values, the results also imply that the important subspace basis with larger singular value is significantly closer to the correctly labeled samples. Thus, the geometric properties, i.e., relation between subspace basis, are still primarily decided by the correctly labeled samples

\textbf{Extreme Learning Scenario}. To study the impacts of low-quality pseudo-label, we evaluate the source model, GOAL w/o warm-up and GOAL w/ warm-up on challenging adaptation dataset DomainNet, where the accuracies are extremely low for some transfer tasks. From the results in Table~\ref{tab:PseudoLabel_DomainNet}, we can observe that the GOAL can still improve the performance of source model. Especially, for some hard tasks, e.g., qdr as source domain, the improvement of applying GOAL model is significantly. Besides, in such an extreme learning scenario, the warm-up training stage is indeed important to improve the quality of preliminary pseudo-labels for geometric principle learning. Quantitatively, the mean accuracy of GOAL w/ warm-up is 33.3\%, which improve the model without warm-up by 6\%. Therefore, these results demonstrate that the GOAL and warm-up strategy are generally effective in extreme learning scenario with low-quality pseudo-labels.

\begin{table}[t]
    \centering
    \caption{Evaluations of different ImageNet pre-trained methods (ResNet-50).}
    \label{tab:pre-trained models}
    \renewcommand{\tabcolsep}{0.35pc} 
    \renewcommand{\arraystretch}{1} 
    \begin{tabular}{cc|cc|ccc}
        \toprule
        \multirow{2}{*}{\textbf{Dataset}} & \multirow{2}{*}{\textbf{Method}} & \multicolumn{2}{c|}{\textbf{w/o FT}} & \multicolumn{3}{c}{\textbf{w/ FT}}  \\
         &  & $\mathcal{L}_{\text{DB}}$ & $\mathcal{L}_{\text{TB}}$ & $\mathcal{L}_{\text{DB}}$ & $\mathcal{L}_{\text{TB}}$ & Acc. \\
        \midrule
        & Cross-Entropy \cite{he2016deep} &  6.2 & 2.5 & 22.5 & 10.2 & 52.3        \\
        VisDA & MoCo (200 epochs) \cite{he2020momentum} & 8.5 & 2.0 & 20.6 & 7.7 & 29.8        \\
        2017 & MoCo v2 (200 epochs) \cite{chen2020improved} &  8.4 & 2.5 & 20.7 & 8.5 & 50.4        \\
        & MoCo v2 (800 epochs) \cite{chen2020improved} &  8.8 & 2.7 & 21.7 & 9.0 & 50.1        \\
        \midrule
        & Cross-Entropy \cite{he2016deep} &  0.9 & 0.2 & 9.2 & 3.5 & 76.3 \\
        Office & MoCo (200 epochs) \cite{he2020momentum} & 1.2 & 0.1 & 9.9 & 2.6 & 50.3        \\
        31 & MoCo v2 (200 epochs) \cite{chen2020improved} &  0.9 & 0.1 & 11.4 & 3.5 & 66.6        \\
        & MoCo v2 (800 epochs) \cite{chen2020improved} & 0.9 & 0.2 & 9.5 & 3.0 & 71.3   \\
        \midrule
        & Cross-Entropy \cite{he2016deep} & 1.4 & 0.4 & 11.0 & 6.5 & 80.8        \\
        Image & MoCo (200 epochs) \cite{he2020momentum} & 1.9 & 0.5 & 12.9 & 6.5 & 67.4        \\
        CLEF & MoCo v2 (200 epochs) \cite{chen2020improved} &  1.4 & 0.5 & 12.0 & 6.8 & 78.5        \\
        & MoCo v2 (800 epochs) \cite{chen2020improved} &  1.3 & 0.5 & 10.3 & 6.2 & 80.3        \\
        \midrule
        & Cross-Entropy \cite{he2016deep} &  1.2 & 0.2 & 11.8 & 3.3 & 53.7        \\
        Office & MoCo (200 epochs) \cite{he2020momentum} & 1.9 & 0.2 & 23.4 & 5.5 & 32.5        \\
        Home & MoCo v2 (200 epochs) \cite{chen2020improved} & 1.5 & 0.2 & 17.3 & 4.2 & 42.5  \\
        & MoCo v2 (800 epochs) \cite{chen2020improved} & 1.5 & 0.2 & 16.8 & 4.4 & 46.6        \\
        \bottomrule
    \end{tabular}
\end{table}

\begin{figure}[t]
    \centering
    \includegraphics[width=0.93\linewidth]{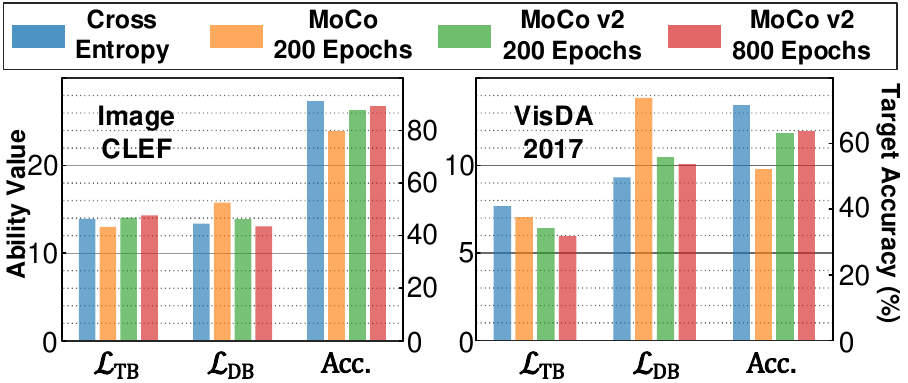}
    \caption{Analysis of different pre-training methods based on transferability value $\mathcal{L}_{\text{TB}}$, discriminability value $\mathcal{L}_{\text{DB}}$ and target accuracy (Acc.).}
    \label{fig:pretraining_method}
\end{figure}

\textbf{Pre-training Methods}.
Existing domain adaptation methods usually initialized the networks with supervised pre-trained parameters on ImageNet \cite{he2016deep}, while recent advances \cite{he2020momentum,chen2020improved} also show the potentials of unsupervised pre-training for downstream applications. To provide insights from the geometric aspects, we conduct experiment to evaluate the transferability and discriminability of different pre-training methods. From the experiment results in Table~\ref{tab:pre-trained models}, we can conclude following observations: 1) for the original representations of pre-training methods, i.e., w/o fine-tuning (FT), the discriminability and transferability of unsupervised representation learning are competitive compared with the supervised pre-training counterpart; 2) for the fine-tuned representations on source domain, i.e., w/ FT, the unsupervised pre-trained model can be adapted to downstream data as efficient as supervised pre-trained model, where the abilities of representations are also similar. Besides, we also compare the performance of GOAL under different pre-trained models, The results in Figure~\ref{fig:pretraining_method} imply that the GOAL with MoCo V2 can also achieve nearly SOTA performance, i.e., about 1.7\% lower than supervised pre-training. In conclusion, these results demonstrate that advanced unsupervised representation pre-training can also provide a good initialization for downstream application with favorable abilities and competitive performance.

\begin{figure}[t]
    \centering
    \includegraphics[width=0.93\linewidth]{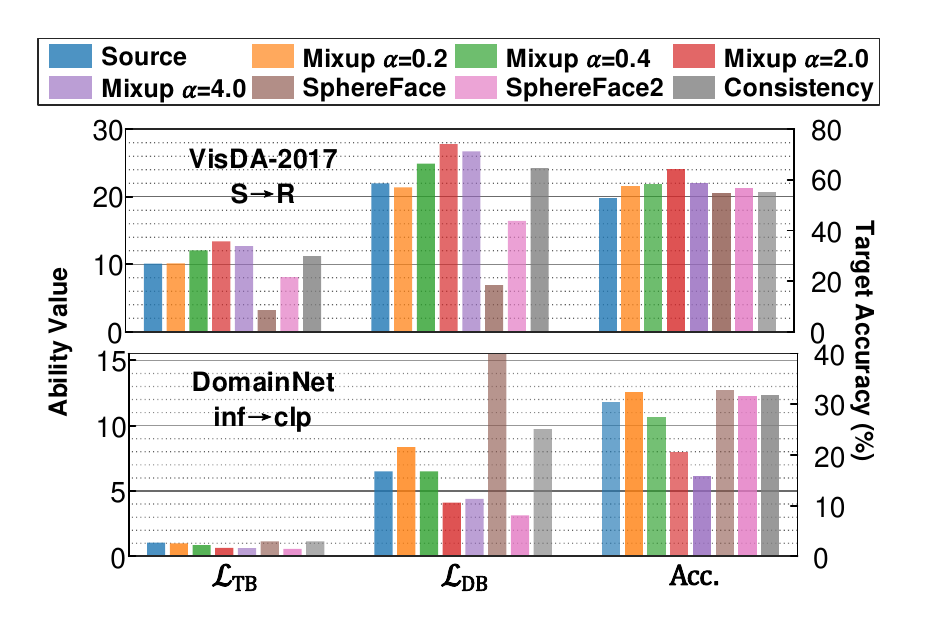} 
    \caption{Analysis of different fine-tuning strategies based on transferability value $\mathcal{L}_{\text{TB}}$, discriminability value $\mathcal{L}_{\text{DB}}$ and target accuracy (Acc.).}
    \label{fig:finetuning_strategies}
\end{figure}

\textbf{Fine-tuning Strategies}. With the proposed geometric principles, we can study the transferability and discriminability of different fine-tuning strategies, which will provide insights into finding advanced training procedure. Thus, we evaluate the source model, source mixup training \cite{zhang2018mixup}, source SphereFace training \cite{liu2017sphereface}, SphereFace2 training \cite{wen2022sphereface2} and source consistency training \cite{hendrycks2020augmix} on large-scale datasets, i.e., VisDA-2017 and DomainNet. Specifically, we apply these strategies on source data to obtain a fine-tuned model. Then the learned representations are extracted to compute the values of geometric principles $\mathcal{L}_{\text{DB}}$ and $\mathcal{L}_{\text{TB}}$, which are considered as the discriminability and transferability of fine-tuned models. As shown in Figure~\ref{fig:finetuning_strategies}, the transferability or discriminability indeed can be enhanced by the advanced fine-tuning strategies, and the target accuracies are generally improved compared with the source model. Specifically, for VisDA-2017 with few classes ($k$=12), the mixup training ensures the highest transferability value and task accuracy. For DomainNet with much more classes ($k$=345), the mixup training with linear interpolation fails to characterize the complex cluster boundary, while the large angular margin learning (i.e., SphereFace and SphereFace2) and robustness learning (i.e., Consistency) are relatively effective. Overall, we can conclude that 1) advanced training strategies are generally superior to the source risk minimization; 2) the mixup training is relatively effective in improving transferability; 3) the angular margin learning is relatively effective in improving discriminability; 4) the consistency learning is generally effective in different scenarios.

\begin{table}[t]
    \centering
    \caption{Ablation study: classification accuracies on target domain.}
    \label{tab:ablation}
    \renewcommand{\tabcolsep}{0.25pc} 
    \begin{tabular}{cccc|cccc}
    \toprule
    \multicolumn{3}{c}{Objective} & Warm & \multirow{2}{*}{\textbf{Image-CLEF}} & \multirow{2}{*}{\textbf{Office-31}} & \multirow{2}{*}{\textbf{Office-Home}} &  \multirow{2}{*}{\textbf{Avg.}} \\
    %\cline{1-2}
     $\mathcal{L}_{\text{TB}}$             & $\mathcal{L}_{\text{DB}}$ & $\mathcal{L}^t_{\text{E}}$  &     Up    &                               &                           &                              &  \\
    \midrule
     $\times$              & \checkmark  & \checkmark & \checkmark          &          90.6                     &        91.1                   &                 65.7             & 82.5  \\
     \checkmark            & $\times$    & \checkmark & \checkmark         &          87.7                     &           90.6                &                64.2              & 80.8 \\
     \checkmark  & \checkmark     & $\times$  & \checkmark &   91.2             &       92.6                    &             68.1                 & 84.0   \\
     \checkmark  & \checkmark     & \checkmark & $\times$ &    90.9             &       93.1                    &             68.2                 & 84.1   \\
    \checkmark &\checkmark              & \checkmark  & \checkmark           &         \textbf{91.3}                      &       \textbf{93.3}                   &             \textbf{68.6}         & \textbf{84.4}  \\
    \bottomrule
    \end{tabular}
\end{table}

\textbf{Ablation Study}. We conduct ablation experiment to analyze the warm-up strategy and the proposed learning principles, i.e., domain equivalence $\mathcal{L}_{\text{TB}}$ and class orthogonality $\mathcal{L}_{\text{DB}}$. The model without warm-up implies that the initialized model is directly optimized with GOAL learning stage in Algorithm~\ref{alg:GOALforUDA}. The classification accuracies are reported in Table~\ref{tab:ablation}. The 1\ts{st} and 2\ts{nd} rows show that the proposed principles improve performance significantly. Specifically, class orthogonality $\mathcal{L}_{\text{DB}}$ is more important when the domain divergence is small, e.g., Image-CLEF. However, when domain divergence becomes larger, the domain equivalence $\mathcal{L}_{\text{TB}}$ is necessary to achieve the SOTA performance. The 3\ts{rd} and 4\ts{th} rows show that warm-up strategy is generally helpful while the entropy objective $\mathcal{L}_{\text{E}}^t$ is significant when there are more classes and larger prediction uncertainty, i.e., Office-31 and Office-Home. Besides, the model without $\mathcal{L}_{\text{E}}^t$ or warm-up still achieves SOTA performance, which verifies that the proposed principles are generally effective. In the last row, the full model is best on all datasets, which demonstrates that $\mathcal{L}_{\text{TB}}$ and $\mathcal{L}_{\text{DB}}$ are consistent and can benefit from others.

\begin{figure}[t]
    \centering
    \includegraphics[width=0.93\linewidth]{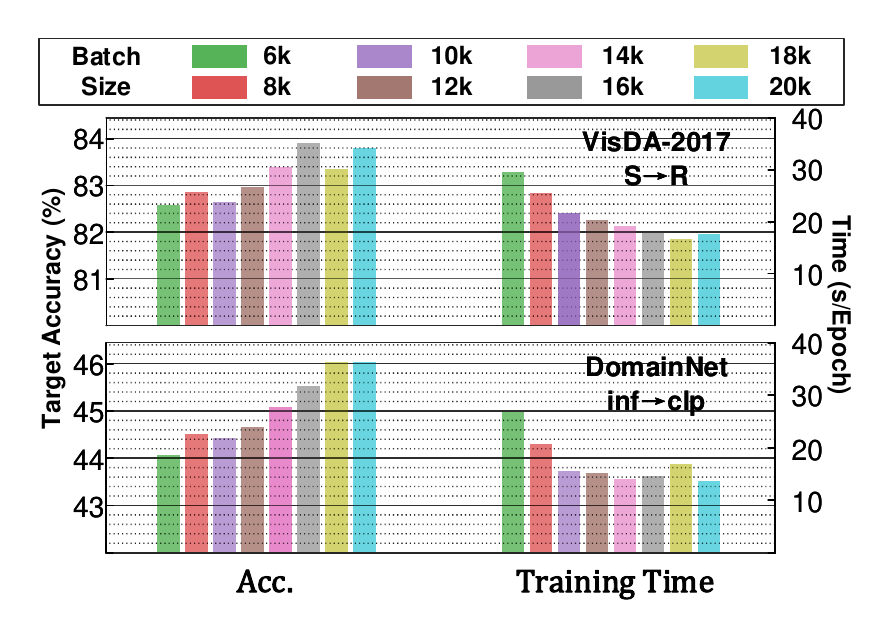} 
    \caption{Evaluations of target accuracy (Acc.) and training time cost (seconds per epoch) under different batch-sizes.}
    \label{fig:BatchSize_comparison}
\end{figure}

\textbf{Training Batch-size}. 
Since the overall model is trained with batch gradient descent, it is also necessary to evaluate the model's performance under different batch-size settings. Based on the designed training procedure, we conduct experiments on VisDA-2017 and DomainNet by varying batch-size from 6k to 20k. The results in Figure~\ref{fig:BatchSize_comparison} provide two key observations. 1) Larger batch-size has higher accuracy, since the geometric learning with larger batch-size ensures a better approximation for intrinsic structure of whole data. 2) Since the complexity of task learning loss and geometric principles are linear w.r.t. the sample-size, the training costs are almost the same when batch-size is large. Besides, the results show that small batch-size may has higher time cost, e.g., 6k and 8k, the main reason is that the small batch-size implies more batches within an epoch, where additional time for more operations (e.g., loading data, inner-iteration for numerical SVD) are induced. Thus, a larger batch-size for empirical setting is preferable.

\section{Conclusion}\label{sec:conclusion}

In this work, we present a systematic study on the transferability and discriminability from the perspective of geometry. The main theoretical results show the \textit{co-regularization relation} between these two aspects and prove the \textit{possibility of learning them simultaneously}. Based on the theoretical results, a feasible range of the regularization parameters for learning balanced abilities is deduced, and an optimization-friendly model is proposed based on the derived learning principles. The theory-driven model ensures the interpretability of the learned invariant representations. The theoretical results are validated by extensive experiments, where the results imply that our model does learn the desired geometric properties.

An interesting future direction can be the learnability of $g$ with a given function class, which connects transferability and discriminability with statistical learning theory. Besides, the geometric-aware principle can be considered for other problems, e.g., multi-modal learning and meta learning.

\bibliography{ref_GOAL}
\bibliographystyle{IEEEtran}

\vspace{-15pt}
\begin{IEEEbiography}[{\resizebox{1.0in}{1.3in}{\includegraphics*{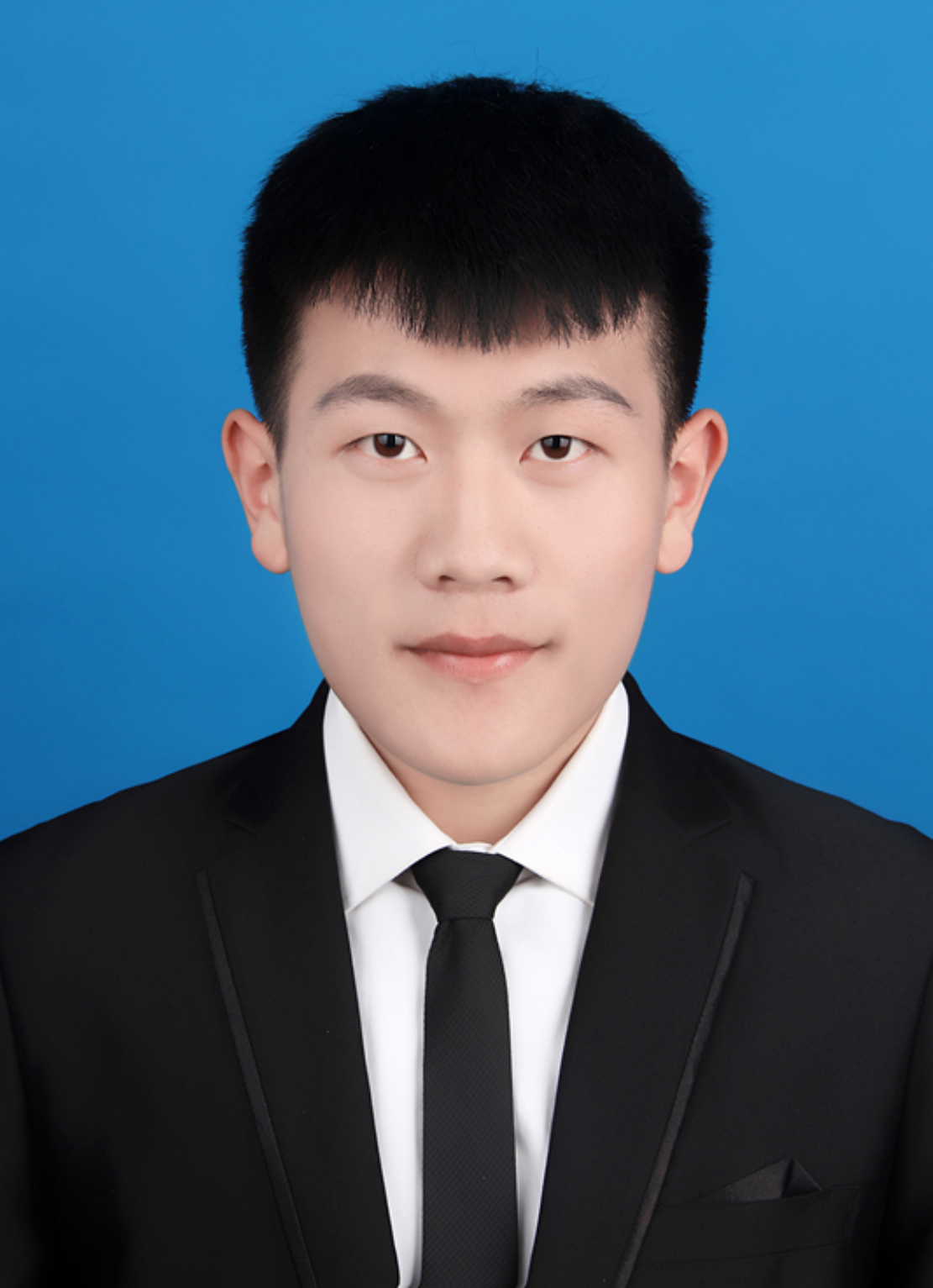}}}]{You-Wei Luo}
received the B.S. degree in Statistics from China University of Mining and Technology, Xuzhou, China, in 2018 and the Ph.D. degree in Mathematics from Sun Yat-sen University, Guangzhou, China, in 2023. He is currently a post-doctoral fellow with the School of Mathematics, Sun Yat-sen University, Guangzhou, China. His research interests include image processing, manifold learning and transfer learning.
\end{IEEEbiography}
    
\vspace{-25pt}
    
\begin{IEEEbiography}[{\resizebox{1.0in}{1.3in}{\includegraphics*{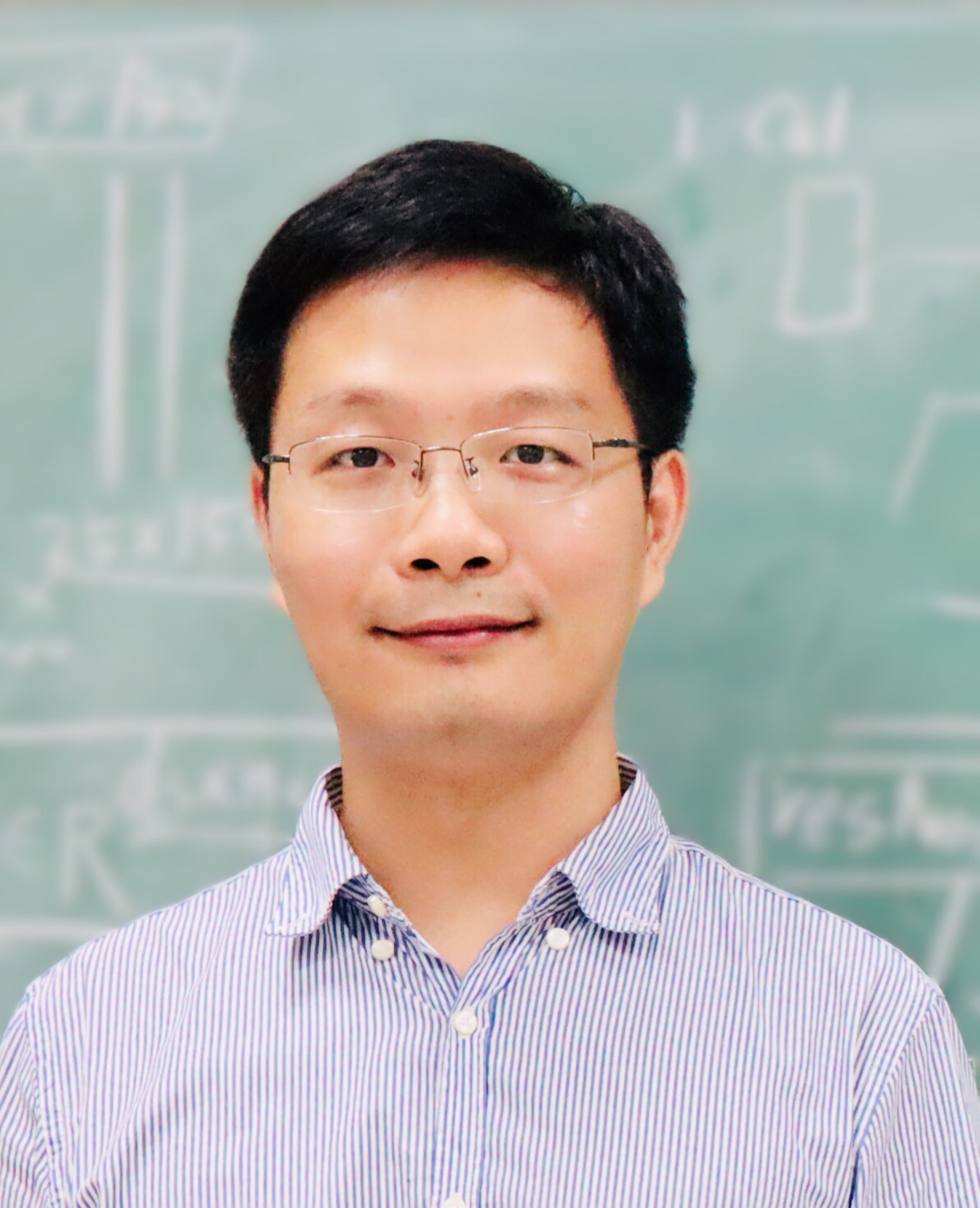}}}]{Chuan-Xian Ren}
received the PhD degree in mathematics from Sun Yat-Sen University, Guangzhou,China, in 2010. During 2010 and 2011, he was with the Department of Electronic Engineering, City University of Hong Kong, as a senior research associate. He is currently a professor with the School of Mathematics, SunYat-SenUniversity.His research interests focus on interdisciplinary study with visual pattern analysis, machine learning and mathematics.
\end{IEEEbiography}

\vspace{-25pt}
\begin{IEEEbiography}[{\resizebox{1.0in}{1.3in}{\includegraphics*{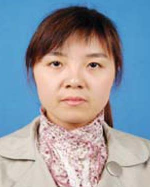}}}]{Xiao-Lin Xu}
received the PhD degree in statistics from Anhui University, Hefei, China, in 2017. She is currently an assistant professor with the School of Statistics and  Mathematics,Guangdong University of Finance and Economics, Guangzhou, China. Her current research interests include statistical learning, modeling and optimization.
\end{IEEEbiography}

\vspace{-25pt}
\begin{IEEEbiography}[{\resizebox{1.0in}{1.3in}{\includegraphics*{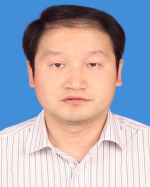}}}]{Qingshan Liu} (Senior Member, IEEE)
received the MS degree from Southeast University, Nanjing, China, in 2000 and the PhD degree from the Chinese Academy of Sciences, Beijing, China, in 2003. He is currently a professor with the School of Computer Science, Nanjing University of Information Science and Technology, Nanjing, China. His research interests include image and vision analysis, machine learning, etc.
\end{IEEEbiography}

\end{document}